\documentclass[sigconf]{acmart}

\usepackage{graphicx}
\usepackage{booktabs}
\usepackage{algorithm}
\usepackage{algpseudocode}
\usepackage{textcomp}
\usepackage{xcolor}
\usepackage{multicol}
\usepackage{multirow}
\usepackage{epstopdf}
\DeclareGraphicsExtensions{.eps}
\usepackage[bottom]{footmisc}
\usepackage{comment}
\usepackage{diagbox}
\usepackage{bm}

\usepackage{subcaption}

\newcommand{\ie}{\textit{i}.\textit{e}.,}
\newcommand{\eg}{\textit{e}.\textit{g}.,}
\newcommand{\etc}{\textit{etc}.}

\usepackage{wrapfig}

\usepackage{listings}
\usepackage{xcolor}

\definecolor{codegreen}{rgb}{0,0.6,0}
\definecolor{codegray}{rgb}{0.5,0.5,0.5}
\definecolor{codepurple}{rgb}{0.58,0,0.82}
\definecolor{backcolour}{rgb}{0.95,0.95,0.92}

\lstdefinestyle{mystyle}{
    backgroundcolor=\color{backcolour},   
    commentstyle=\color{codegreen},
    keywordstyle=\color{magenta},
    numberstyle=\tiny\color{codegray},
    stringstyle=\color{codepurple},
    basicstyle=\ttfamily\footnotesize,
    breakatwhitespace=false,         
    breaklines=true,                 
    captionpos=b,                    
    keepspaces=true,                 
    showspaces=false,                
    showstringspaces=false,
    showtabs=false,                  
    tabsize=2
}

\AtBeginDocument{%
  \providecommand\BibTeX{{%
    \normalfont B\kern-0.5em{\scshape i\kern-0.25em b}\kern-0.8em\TeX}}}

\copyrightyear{2024}
\acmYear{2024}
\setcopyright{acmlicensed}\acmConference[KDD'24]{Proceedings of the 30th
ACM SIGKDD Conference on Knowledge Discovery and Data Mining}{August
25--29, 2024}{Barcelona, Spain}
\acmBooktitle{Proceedings of the 30th ACM SIGKDD Conference on Knowledge
Discovery and Data Mining (KDD'24), August 25--29, 2024, Barcelona, Spain}
\acmDOI{10.1145/3637528.3671536}
\acmISBN{979-8-4007-0490-1/24/08}

\begin{document}

\title{An Open and Large-Scale Dataset for Multi-Modal  Climate Change-aware Crop Yield Predictions}

\author{Fudong Lin}
\affiliation{%
  \institution{University of Delaware}
  \city{Newark}
  \state{DE}
  \country{USA}
}

\author{Kaleb Guillot}
\affiliation{
  \institution{University of Louisiana at Lafeyette}
  \city{Lafeyette}
  \state{LA}
  \country{USA}}

\author{Summer Crawford}
\affiliation{
  \institution{University of Louisiana at Lafeyette}
  \city{Lafeyette}
  \state{LA}
\country{USA}
}

\author{Yihe Zhang}
\affiliation{
  \institution{University of Louisiana at Lafeyette}
  \city{Lafeyette}
  \state{LA}
  \country{USA}
  }

  \author{Xu Yuan}
  \authornote{Corresponding author: Dr. Xu Yuan (xyuan@udel.edu)}
  \affiliation{%
    \institution{University of Delaware}
    \city{Newark}
    \state{DE}
    \country{USA}
  }

\author{Nian-Feng Tzeng} 
\affiliation{
  \institution{University of Louisiana at Lafeyette}
  \city{Lafeyette}
  \state{LA}
  \country{USA}
}

\renewcommand{\shortauthors}{Fudong and Kaleb, et al.}

\begin{abstract}
  Precise crop yield predictions are of national importance for ensuring food security and sustainable agricultural practices. 
  While \textit{AI-for-science} approaches have exhibited promising achievements
  in solving many scientific problems such as drug discovery, precipitation nowcasting, \etc,
  the development of deep learning models for predicting crop yields 
  is constantly hindered by the lack of an open and large-scale deep learning-ready dataset with multiple modalities to accommodate sufficient information.
  To remedy this, we introduce the CropNet dataset, 
  the first \textit{terabyte-sized}, publicly available, and multi-modal dataset 
  specifically targeting climate change-aware crop yield predictions for the contiguous United States (U.S.) continent at the county level.
  Our CropNet dataset is composed of three modalities of data, \ie\ Sentinel-2 Imagery, WRF-HRRR Computed Dataset, and USDA Crop Dataset,
  for over $2200$ U.S. counties spanning $6$ years (2017-2022),
  expected to facilitate researchers in developing versatile  deep learning models for timely and precisely predicting crop yields at the county-level,
  by accounting for the effects of  both short-term growing season weather variations 
  and long-term climate change on crop yields.
  Besides, we develop the CropNet package,
  offering three types of APIs,
  for facilitating researchers in
  downloading the CropNet data on the fly over the time and region of interest,
  and flexibly building their deep learning models for accurate crop yield predictions. 
  Extensive experiments have been conducted on our CropNet dataset via
  employing various types of deep learning solutions,
  with the results validating the general applicability and the efficacy  of the CropNet dataset 
  in climate change-aware crop yield predictions.
  We have officially released our CropNet dataset on Hugging Face Datasets
  \textcolor{magenta}{{https://huggingface.co/datasets/CropNet/CropNet}}
  and our CropNet package on the Python Package Index (PyPI) 
  \textcolor{magenta}{\url{https://pypi.org/project/cropnet}}.
  Code and tutorials are available at \textcolor{magenta}{\url{https://github.com/fudong03/CropNet}}.
\end{abstract}




\maketitle

\section{Introduction}
\label{sec:intro}

\begin{figure*} [!t] 
  \centering
  \includegraphics[width=.90
  \textwidth]
  {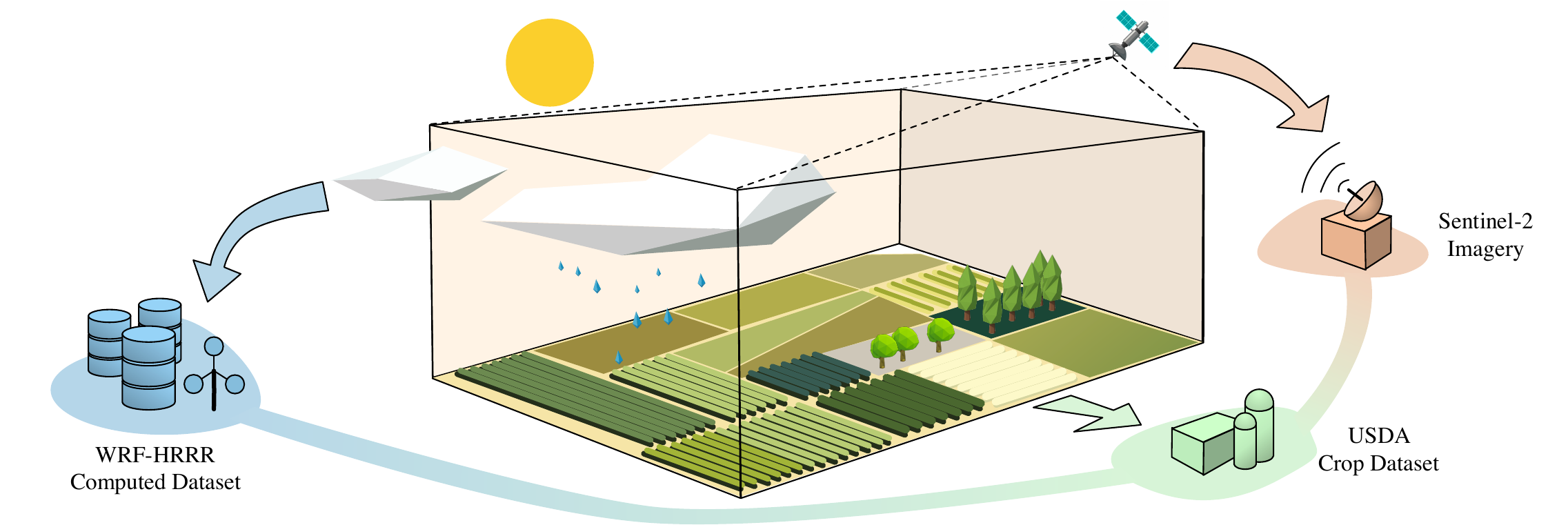}
  \vspace{-0.5 em}
  \caption{
    Our CropNet dataset is composed of three modalities of data,
    \ie\ Sentinel-2 Imagery, WRF-HRRR Computed Dataset, and USDA Crop Dataset,
    providing satellite images, meteorological parameters, and county-level crop yield information, respectively.
  }
  \label{fig:dataset-multiple-modalities}
  \vspace{-1 em}
\end{figure*}

Precise crop yield prediction is essential for 
early agricultural planning~\cite{khaki2021simultaneous},
timely management policy adjustment~\cite{turchetta:nips22:crop_management},
informed financial decision making~\cite{ansarifar2021interaction},
and national food security~\cite{mourtzinis2021advancing}.
Recent advancements in deep neural networks (DNNs) have achieved impressive performance across various domains~\cite{alex:nips12:alex_net,he:cvpr16:resnet,vaswani:nips17:attention,dosovitskiy:iclr21:vit,he:ijcai21:gan,wang2022learning,chen2022relax,lin:cikm22:vae,mo2022cvlight,chen2022explain,ma:nips22:healthcare,feng2023robotic,lin:ecml_pkdd23:vit,ruan:iclr23:causal,li2024feature,mo202424llm,zhang2024large,zhan2024meta,song2024bundledslam}. 
Building upon these advancements, plenty of studies have employed spatial-temporal DNNs~\cite{zhan2019adaptive,wang2022novel,ma2023eliminating,li2023stock,liu2023financial,wang2023st_gin,li2022detecting,ma2024data,mo2024pi_neugode,li2022automated,he:kdd22:st} 
to predict crop yields with increased timeliness and precision~\cite{khaki2020cnn, khaki2021simultaneous,garnot:iccv21:panoptic,wu2021spatiotemporal,cheng2022high,fan:aaai23:crop_prediction,fudong:iccv23:mmst_vit}.
However, they often applied their personally curated and limit-sized datasets, with somewhat mediocre prediction performance. 
There is an urgent need for new large-scale and deep learning-ready datasets tailored specifically for wide use in crop yield predictions.

Recently, some studies
\cite{veillette:nips20:sevir,garnot:iccv21:panoptic,tseng:nips21:crop_harvest,cornebise:nipc22:world_strat,he2020curvanet,chen:nips22:rain_net,ashkboos:nips22:ens10,li2023understanding,cottam2024large,kangrui:icassp24:s2e}
have developed open and large-scale satellite imagery (or meteorological parameter) datasets,
flexible for being adopted to agricultural-related tasks,
\eg\ crop type classification~\cite{tseng:nips21:crop_harvest}.
Unfortunately, two limitations impede us from applying them directly to crop yield predictions in general.
First, they lack ground-truth crop yield information, 
making them unsuitable for crop yield predictions.
Second, they provide only one modality of data (\ie\ either satellite images or meteorological parameters),
while accurate crop yield predictions often need to track the crop growth 
and capture the meteorological weather variation effects on crop yields simultaneously,
calling for multiple modalities of data.
To date, the development of  a large-scale dataset with multiple modalities,
targeting specifically for county-level crop yield predictions remains open and challenging.

\begin{table}   [htbp]  
    \scriptsize
    \centering
    \setlength\tabcolsep{10 pt}
    \caption{
        Dataset comparison
        }
        \vspace{-1.0 em}
        \begin{tabular}{@{}ccc@{}}
            \toprule
            Dataset    & Size (GB) & Data Modality         \\ \midrule
            SEVIR~\cite{veillette:nips20:sevir}      & 970      & satellite imagery \\
            DENETHOR~\cite{lukas:nips21:denthor}     & 254       & satellite imagery \\
            PASTIS~\cite{garnot:iccv21:panoptic}     & 29       & satellite imagery \\
            WorldStrat~\cite{cornebise:nipc22:world_strat} & 107      & satellite imagery \\ 
            RainNet~\cite{chen:nips22:rain_net}    & 360      & satellite imagery \\ 
            ENS-10~\cite{ashkboos:nips22:ens10}    & 3072      & meteorological parameters \\ \midrule
            Our CropNet dataset & 2362 & \begin{tabular}[c]{@{}c@{}}satellite imagery\\ meteorological parameters\\ crop information\end{tabular} \\ \bottomrule
        \end{tabular}
    \label{tab:overview-comparison}
    \vspace{-1.0 em}
\end{table}

In this work, we aim to craft such a dataset, called CropNet, 
the first \textit{terabyte-sized} and publicly available dataset 
with multiple modalities,
designed specifically for county-level crop yield predictions
across the United States (U.S.) continent.
As shown in Figure~\ref{fig:dataset-multiple-modalities},
the CropNet dataset is composed of  three modalities of data,
\ie\ Sentinel-2 Imagery, WRF-HRRR Computed Dataset, and USDA Crop Dataset,
covering a total of $2291$ U.S. counties 
from $2017$ to $2022$.
%
In particular, the Sentinel-2 Imagery,
acquired from the Sentinel-2 mission~\cite{sentinel-2}, 
provides two categories of satellite images,
\ie\ agriculture imagery (AG) and normalized difference vegetation index (NDVI),
for precisely monitoring the crop growth on the ground.
The WRF-HRRR Computed Dataset, 
obtained from the WRF-HRRR model~\cite{hrrr},
offers daily and monthly meteorological parameters,
accounting respectively for the short-term weather variations
and the long-term climate change.
The USDA Crop Dataset,
sourced from the USDA Quick Statistic website~\cite{usda},
contains annual crop yield information
for four major crops,
\ie\ corn, cotton, soybean, and winter wheat,
grown on the contiguous U.S. continent,
serving as the ground-truth label for crop yield prediction tasks.
Table~\ref{tab:overview-comparison} summarizes the dataset comparison 
between our CropNet dataset and pertinent datasets.

Since the data in our CropNet dataset are obtained from different data sources,
we propose a novel data alignment solution
to make Sentinel-2 Imagery, WRF-HRRR data, and USDA crop yield data spatially and temporally aligned.
Meanwhile, three modalities of data are stored in carefully designed file formats,
for improving the accessibility, readability, and storage efficiency of our CropNet dataset.
The key advantage of our CropNet dataset is to facilitate researchers
in developing crop yield prediction models that are aware of climate change,
by taking into account the effects of 
(1) the short-term weather variations, governed by daily parameters during the growing season,
and (2) the long-term climate change, governed by monthly historical weather variations, on crop growth.
Furthermore, we have developed the CropNet package, including three types of APIs, 
expected to assist researchers and practitioners in
(1) dynamically downloading the CropNet data based on the specific time and region of interest
and (2) flexibly building climate change-aware deep learning models for accurate crop yield predictions at the county level.

Our experimental results validate that the CropNet dataset can be easily adopted by the prominent deep learning models, 
such as Long Short-Term Memory (LSTM)-based, Convolutional Neural Network (CNN)-based, 
Graph Neural Network~\cite{kipf:iclr17:gnn} (GNN)-based, 
and Vision Transformer~\cite{dosovitskiy:iclr21:vit} (ViT)-based models, 
for timely and precise crop yield predictions.
Additionally, our CropNet dataset demonstrates its versatile applicability to
boost the generalization capabilities
of deep neural networks (DNNs),
thanks to its abundant visual satellite imagery and numerical meteorological data.

\section{Data Sources}
\label{sec:preliminary}

Our CropNet dataset is crafted from three different data sources, as listed below.
\par\smallskip\noindent
{\bf Sentinel-2 Mission.} 
The Sentinel-2 mission~\cite{sentinel-2}, launched in $2015$, serves as an essential earth observation endeavor.
With its $13$ spectral bands and high revisit frequency of $5$ days,
the Sentinel-2 mission provides wide-swath, high-resolution, multi-spectral satellite images
for a wide range of applications, such as climate change, agricultural monitoring, \etc

\par\smallskip\noindent
{\bf WRF-HRRR Model.} 
The High-Resolution Rapid Refresh (HRRR)~\cite{hrrr}
is a Weather Research \& Forecasting Model (WRF)-based forecast modeling system, 
which hourly forecasts weather parameters for the whole United States continent with a spatial resolution of $3$km.
We take the HRRR assimilated results archived in the University of Utah for use,
which provides several crop growth-related parameters, 
\textit{e.g.}, temperature, precipitation, wind speed, relative humidity, radiation, \textit{etc.},
beginning with July 2016.

\par\smallskip\noindent
{\bf USDA.} 
The United States Department of Agriculture (USDA)~\cite{usda}
provides annual crop information for major crops grown in the U.S., 
including corn, cotton, soybeans, wheat, \textit{etc.}, at the county level.
The statistical data include the planted areas, the harvested areas, the production, and the yield for each type of crop, dating back to $1850$ at the earliest.

\begin{figure*}  [!t] 
    \centering
    \includegraphics[width=.80
    \textwidth]{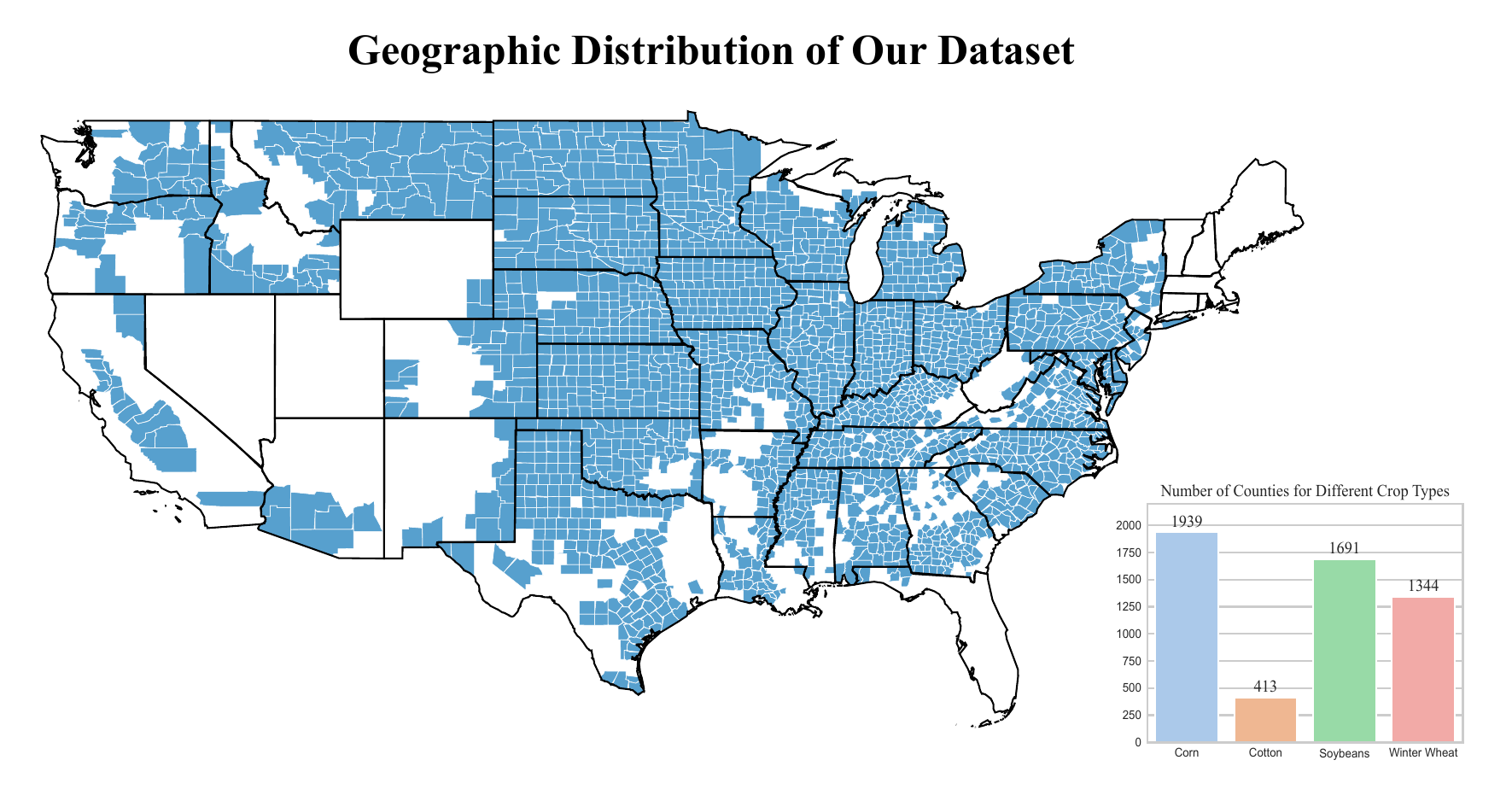}
    \vspace{-1em}
    \caption{
        Geographic distribution of our CropNet dataset across $2291$ U.S. counties. 
        The rightmost bar chart provides the number of counties
        corresponding to each of the four crop types included in the USDA Crop Dataset.
    }
    \label{fig:dataset-geo-info}
    \vspace{-1.0 em}
\end{figure*}

\section{Our CropNet Dateset}
\label{sec:dataset}

%

\subsection{Motivation} 
\label{sec:motivation}

The large-scale data with multiple modalities comprising satellite images, 
numerical meteorological weather data, and crop yield statistic data, are essential for tracking crop growth and correlating the weather variation's effects on crop yields, to be used for timely and precisely predicting crop yields at the county level. 
To date, such an open and large-scale dataset intended for county-level crop yield prediction is still absent.
In this benchmark article, we plan to design and publish such an open and large-scale dataset, called CropNet, 
with multiple modalities, consisting of visual satellite images, numerical meteorological parameters, and crop yield statistic data, across the  U.S. continent.
Notably, not all U.S. counties are suitable for crop planting, 
so our dataset only includes the data 
corresponding to 2291 U.S. counties over 3143 counties in total 
(see Figure~\ref{fig:dataset-geo-info} for its geographic distribution).
Such a multi-modal dataset is valuable for researchers and practitioners to design and test various deep learning models for crop yield predictions, 
by taking into account the effects of both short-term growing season weather variations and long-term climate change on crop yields.


\begin{table*} [!t] 
    \scriptsize
    \centering
    \setlength\tabcolsep{12 pt}
    \caption{
        Overview of our CropNet dataset
        }
        \vspace{-0.5 em}
        \begin{tabular}{@{}ccccc@{}}
            \toprule
            Data Modality               & Size       & Spatial Resolution   & Temporal Resolution     & Content                      \\ \midrule
            Sentinel-2 Imagery          & 2326.7 GB   & 40 m                & 14 days                 & satellite images with 224x224 pixels \\ 
            WRF-HRRR Computed Dataset   & 35.5 GB     & 9x9 km              & 1 day or 1 month        & weather parameters         \\
            USDA Crop Dataset           & 2.3 MB      & county-level        & 1 year                  & crop information            \\ \bottomrule
            \end{tabular}
    \label{tab:dataset-overview}
    \vspace{-1 em}
\end{table*}

\subsection{Overview of Our CropNet Dataset} 
\label{sec:dataset-overview}
\begin{figure*} [!t] 
    \centering
    \captionsetup[subfigure]{justification=centering}
    \begin{subfigure}[t]{0.40\textwidth}
        \centering
        \includegraphics[width=\textwidth]{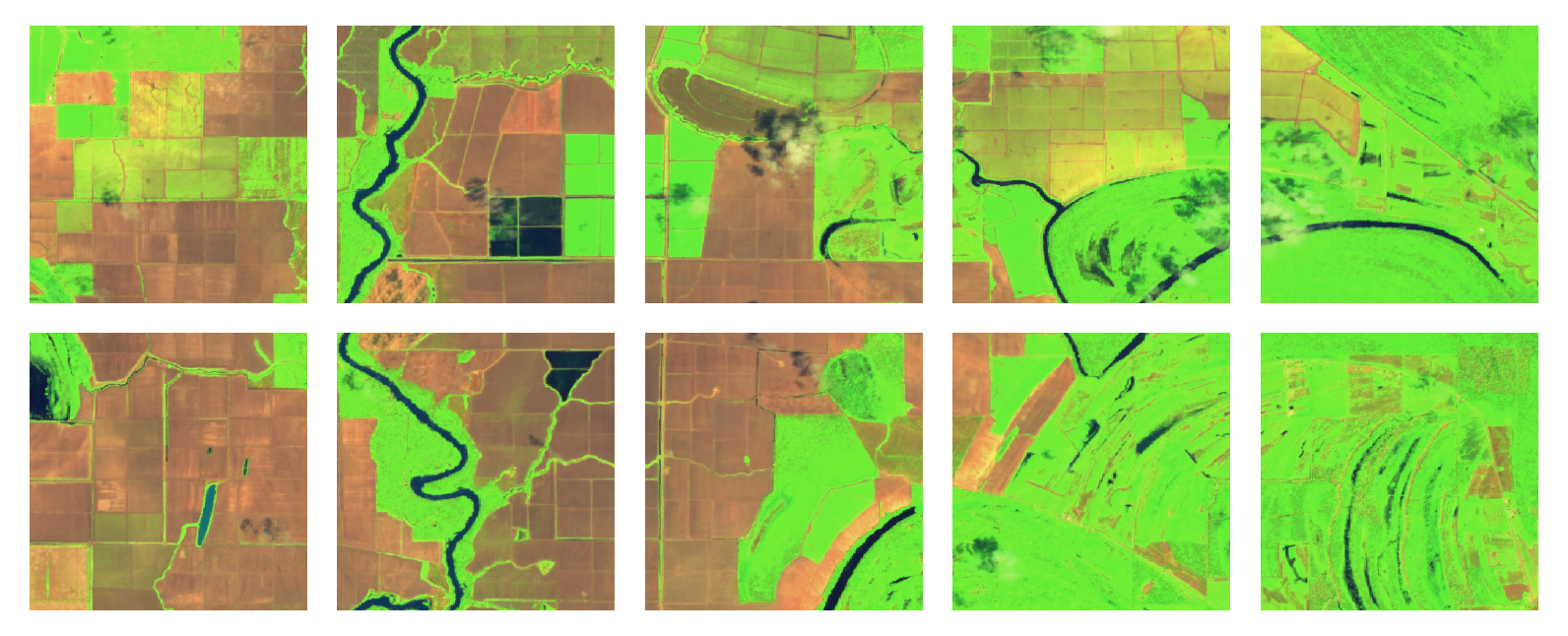}
        \vspace{-1.0 em}
        \caption{AG images in the summer}
        \label{fig:dataset-ag-exp-summer}
    \end{subfigure}
    \quad 
    \begin{subfigure}[t]{0.40\textwidth}
        \centering
        \includegraphics[width=\textwidth]{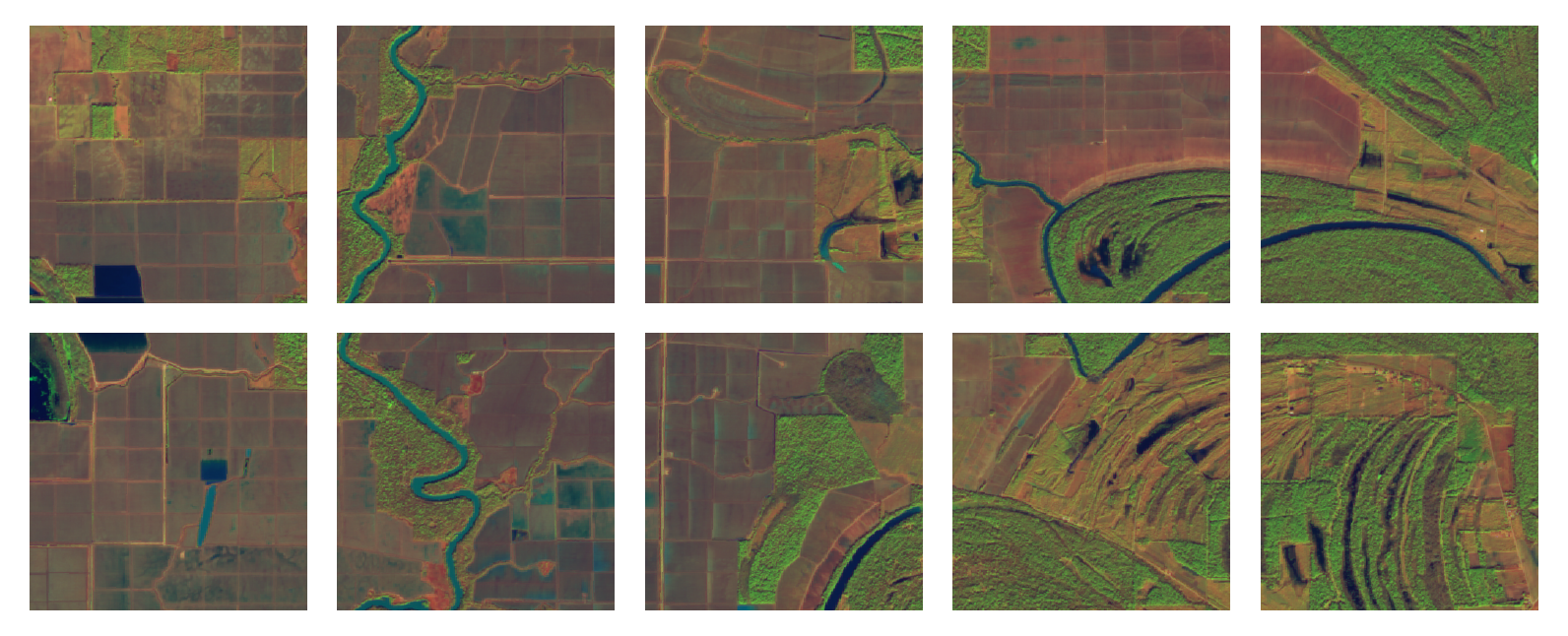}
        \caption{AG images in the winter}
        \label{fig:dataset-ag-exp-winter}
    \end{subfigure}
    
    \begin{subfigure}[t]{0.40\textwidth}
        \centering
        \includegraphics[width=\textwidth]{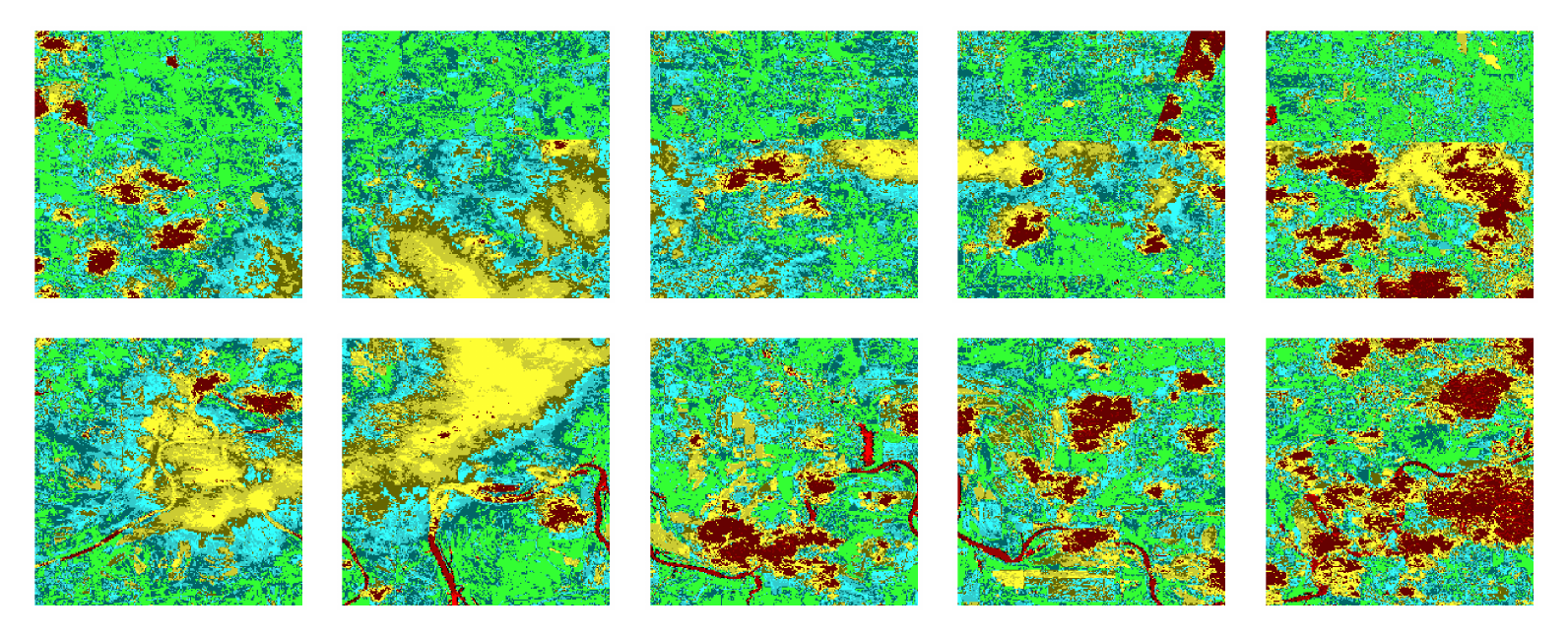}
        \caption{NDVI images in the summer}
        \label{fig:dataset-ndvi-exp-summer}
    \end{subfigure}
    \quad 
    \begin{subfigure}[t]{0.40\textwidth}
        \centering
        \includegraphics[width=\textwidth]{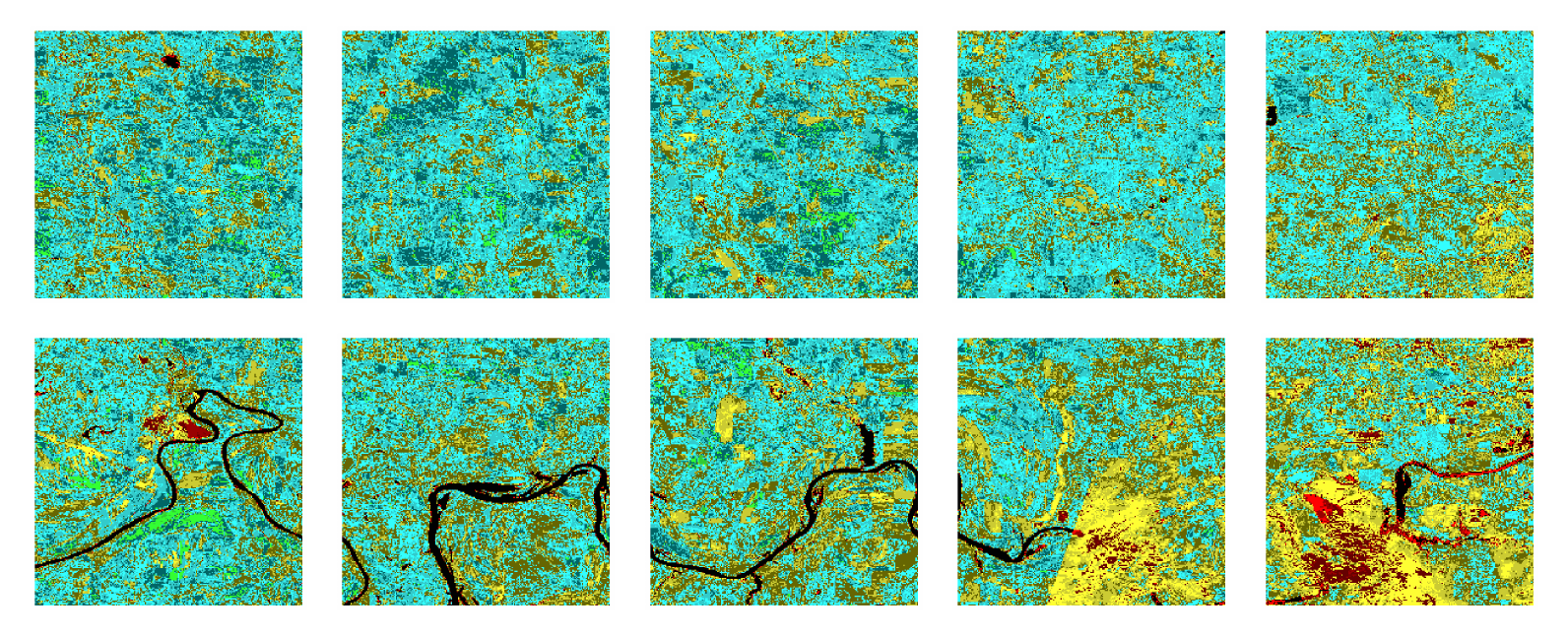}
        \caption{NDVI images in the winter}
        \label{fig:dataset-ndvi-exp-winter}
      \end{subfigure}
    \vspace{-0.8 em}
    \caption{
        Examples of agriculture imagery (AG, see \ref{fig:dataset-ag-exp-summer} 
        and \ref{fig:dataset-ag-exp-winter}) 
        and normalized difference vegetation index (NDVI, see \ref{fig:dataset-ndvi-exp-summer} 
        and \ref{fig:dataset-ndvi-exp-winter}) in Sentinel-2 Imagery.
    }
    \label{fig:dataset-sentinel2-exp}
  \end{figure*}

\begin{figure*} [!t] 
    \centering
    \captionsetup[subfigure]{justification=centering}
    \begin{subfigure}[t]{0.40\textwidth}
        \centering
        \includegraphics[width=\textwidth]{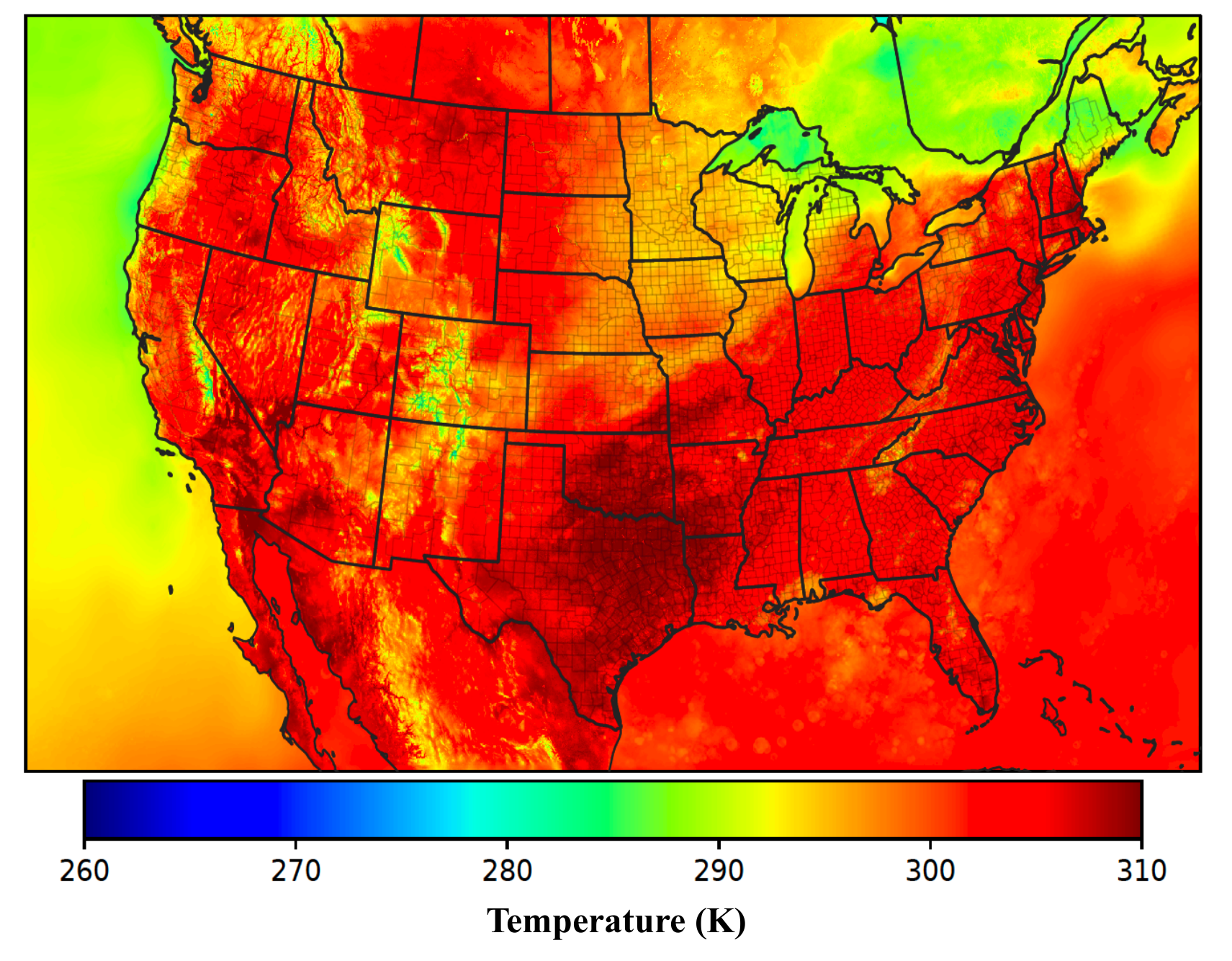}
        \caption{Temperature in the summer}
        \label{fig:dataset-hrrr-tem-summer-m}
    \end{subfigure}
    \quad 
    \begin{subfigure}[t]{0.40\textwidth}
        \centering
        \includegraphics[width=\textwidth]{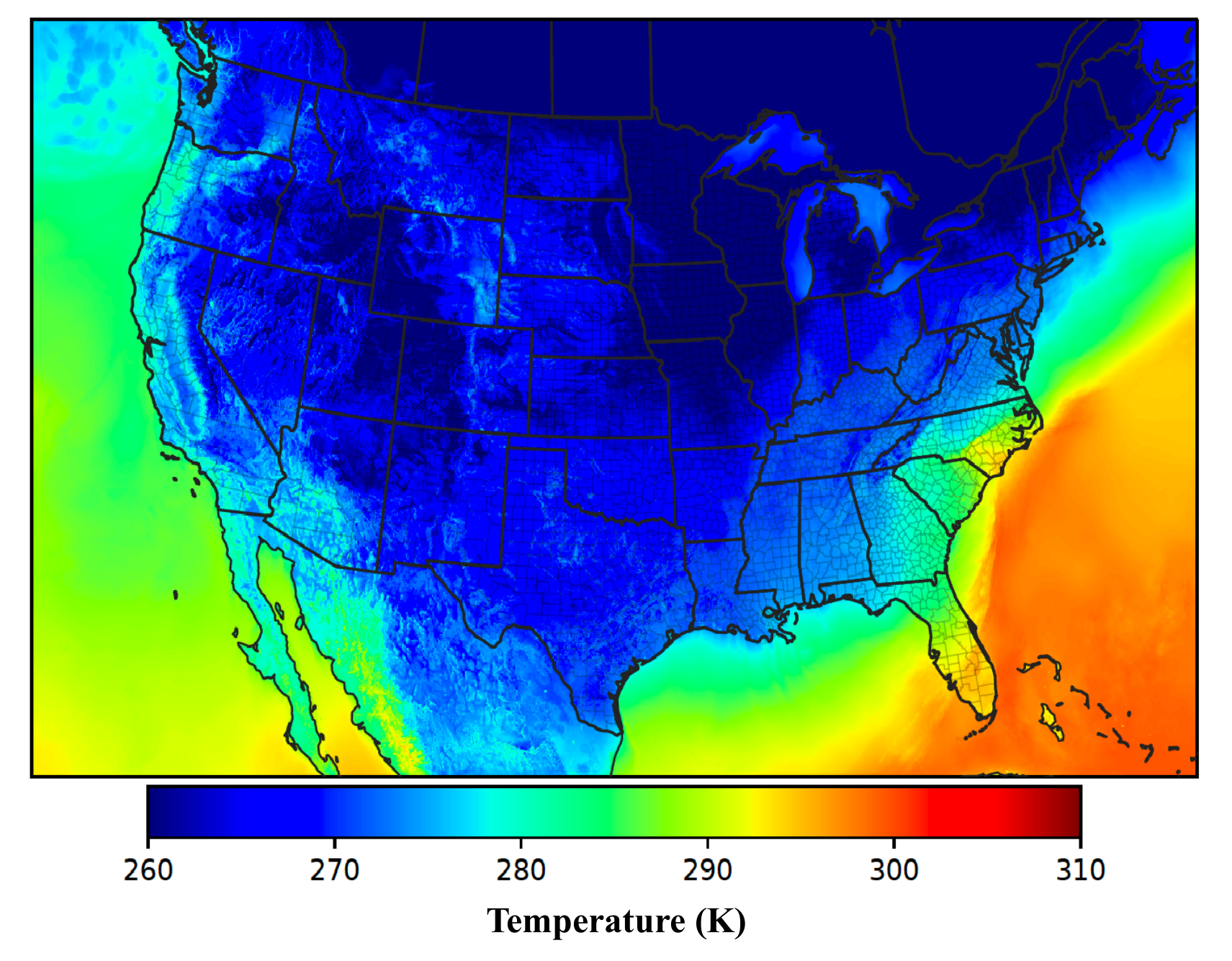}
        \caption{Temperature in the winter}
        \label{fig:dataset-hrrr-tem-winter-m}
    \end{subfigure}
    \vspace{-0.5 em}
    \caption{
     Examples of the temperature parameters in the WRF-HRRR Computed Dataset. 
    }
    \label{fig:dataset-hrrr-example}
\end{figure*}

\begin{figure*}  [!t] 
    \centering
    \includegraphics[width=.80
    \textwidth]{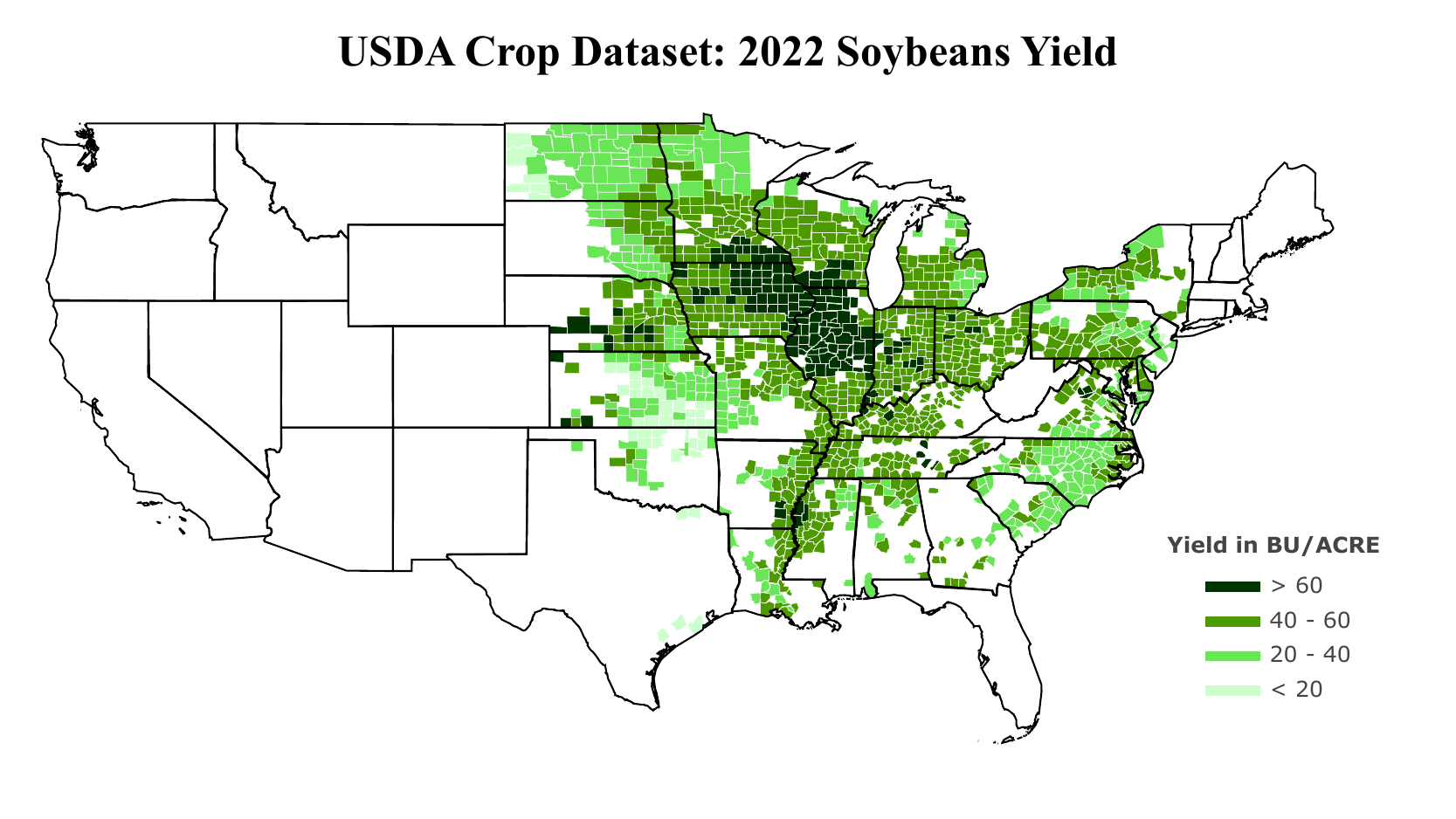}
     \vspace{-0.5 em}
    \caption{
        Illustration of USDA Crop Dataset for 2022 soybeans yields.
    }
    \label{fig:dataset-usda-example}
    \vspace{-1.0 em}
\end{figure*}

Our CropNet dataset is composed of three modalities of data, 
\ie\ Sentinel-2 Imagery, WRF-HRRR Computed Dataset, and USDA Crop Dataset,
spanning from $2017$ to $2022$ (\ie\ $6$ years) across $2291$ U.S. counties.
Figure~\ref{fig:dataset-geo-info} shows the geographic distribution of our dataset.
Since crop planting is highly geography-dependent,
Figure~\ref{fig:dataset-geo-info} also provides the number of counties 
corresponding to each crop type in the USDA Crop Dataset
(see the rightmost bar chart). 
Notably, four of the most popular crops,
\ie\ corn, cotton, soybeans, and winter wheat,
are included in our CropNet dataset,
with satellite imagery and the meteorological data covering all $2291$ counties.
Table~\ref{tab:dataset-overview} overviews our CropNet dataset.
Its total size is $2362.6$ GB, with $2326.7$ GB of visual data for Sentinel-2 Imagery, $35.5$ GB of numerical data for WRF-HRRR Computed Dataset,
and $2.3$ MB of numerical data for USDA Crop Dataset.
Specifically, Sentinel-2 Imagery contains two types of satellite images (\ie\ AG and NDVI), 
both with a spatial resolution of around $40$ meters
(covering an area of $9$x$9$ km with $224$x$224$ pixels)
as well as a revisit frequency of $14$ days.
Figures~\ref{fig:dataset-ag-exp-summer} (or \ref{fig:dataset-ag-exp-winter}) 
and \ref{fig:dataset-ndvi-exp-summer} (or \ref{fig:dataset-ndvi-exp-winter})
respectively depict examples of AG and NDVI images in the summer (or winter).
The WRF-HRRR Computed Dataset provides daily (or monthly) meteorological parameters 
gridded at the spatial resolution of $9$ km in a one-day (or one-month) interval.
Figures~\ref{fig:dataset-hrrr-tem-summer-m} and \ref{fig:dataset-hrrr-tem-winter-m}
visualize the temperature in the WRF-HRRR Computed Dataset 
for the summer and the winter, respectively.
The USDA Dataset offers crop information
for four types of crops each on the county-level basis,
with a temporal resolution of one year.
Figure~\ref{fig:dataset-usda-example} shows
the example for the USDA Crop Dataset,
depicting 2022 soybeans yields across the U.S. continent.
%

\subsection{Data Collection and Preparation} 
\label{sec:data-collection}

\par
\begin{figure*} [!t] 
    \centering
    \captionsetup[subfigure]{justification=centering}
    \begin{subfigure}[t]{0.25\textwidth}
        \centering
        \includegraphics[width=\textwidth]{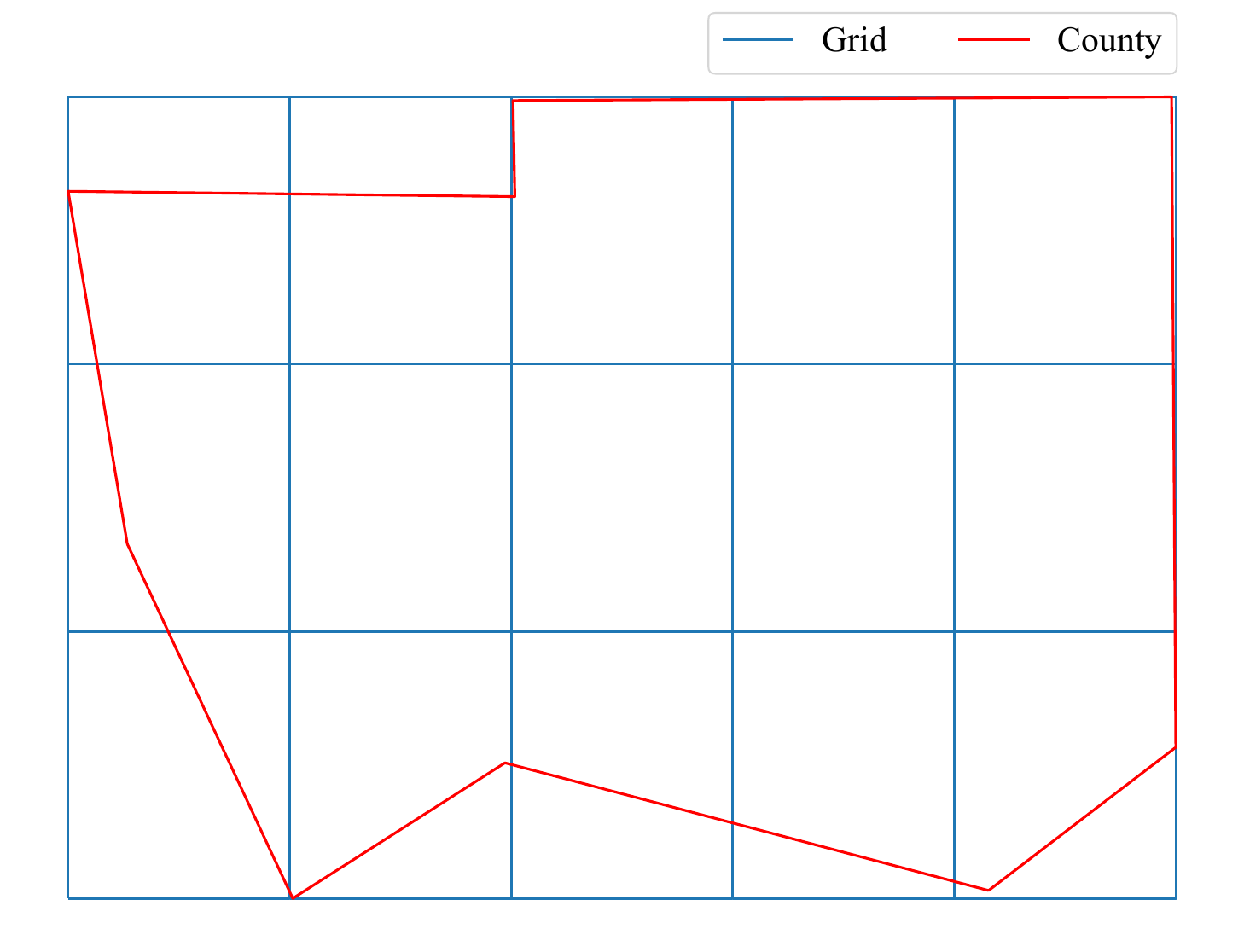}
        \caption{Boundaries}
        \label{fig:dataset-grid-example}
    \end{subfigure}
    \begin{subfigure}[t]{0.30\textwidth}
        \centering
        \includegraphics[width=\textwidth]{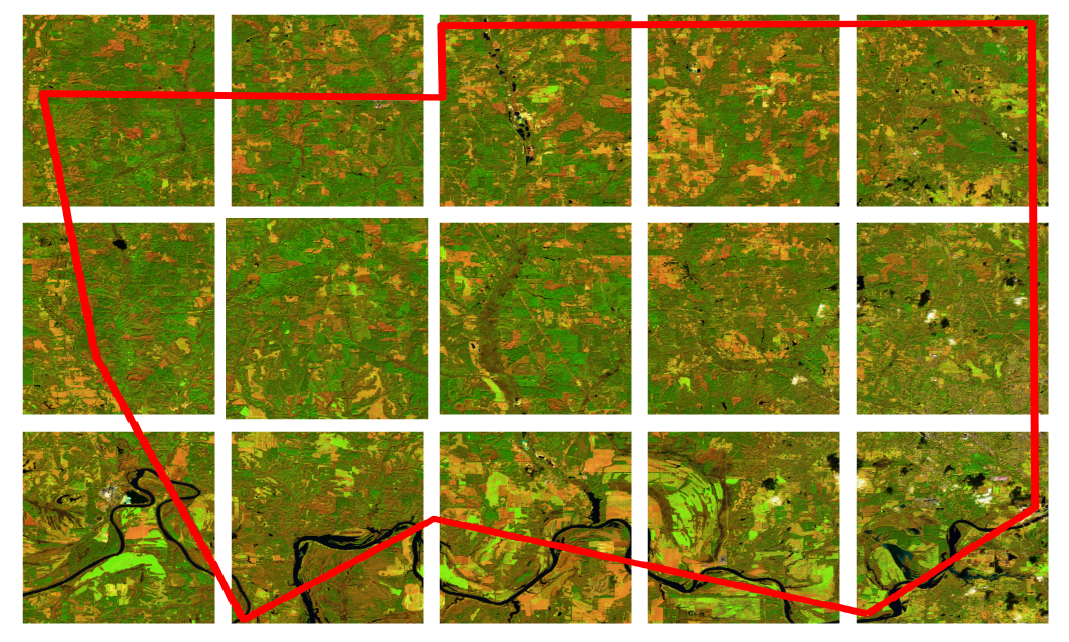}
        \caption{Satellite images}
        \label{fig:dataset-grid-example-ag}
    \end{subfigure}
    \begin{subfigure}[t]{0.37\textwidth}
        \centering
        \includegraphics[width=\textwidth]{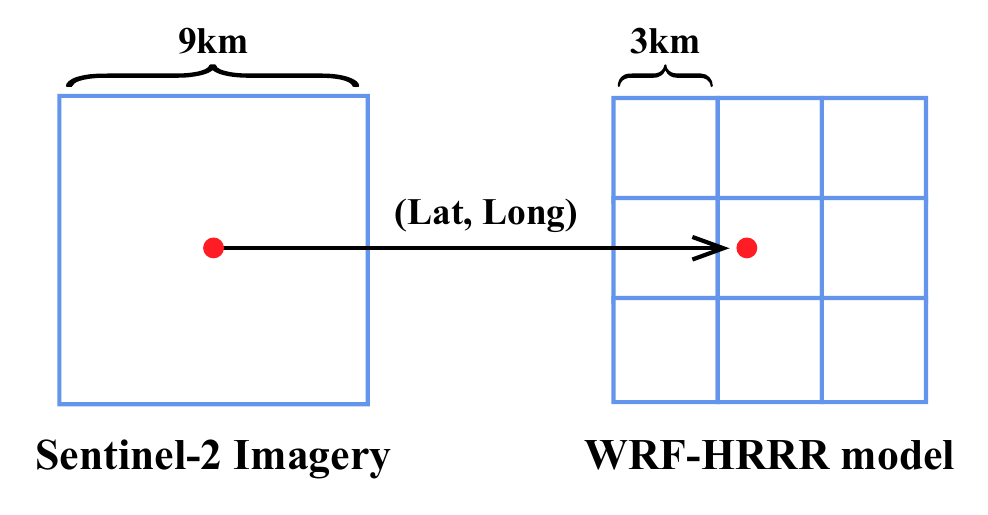}
        \caption{Spatial resolution alignment}
        \label{fig:method-hrrr-grid}
    \end{subfigure}
    \vspace{-0.5 em}
    \caption{
        Illustration of county partitioning 
        (\ie\ \ref{fig:dataset-grid-example} and \ref{fig:dataset-grid-example-ag})
        and spatial resolution alignment (\ie\ \ref{fig:method-hrrr-grid}).
     (a) Boundaries for one county (\ie\ the red line) and 
     the corresponding high-resolution grids (\ie\ the blue line).
     (b) Satellite images in the Sentinel-2 Imagery for representing the county.
     (c) One 3x3km and its surrounding eight grids in the WRF-HRRR model
     are used for aligning with one 9x9km grid in the Sentinel-2 Imagery.
    }
    \label{fig:dataset-county-gird}
    \vspace{-0.5 em}
\end{figure*}

\smallskip\noindent
{\bf Sentinel-2 Imagery.} 
We utilize the Sentinel Hub Processing API~\cite{process-api} to 
acquire satellite images from the Sentinel-2 mission
at a processing level of Sentinel-2 L1C, 
with a maximum allowable cloud coverage of $20 \%$,
three spectral bands (\ie\ B02, B08, and B11) for AG images
and two bands (\ie\ B04 and B08) for NDVI images.
Satellite images are obtained at the revisit frequency of $14$ days instead of the original highest revisit frequency of 5 days. 
The reason is that the 5-day revisit frequency under our cloud coverage setting 
results in a large number of duplicate satellite images, 
according to  our empirical study (refer to Appendix~\ref{sup:sentinel-2-setting} 
for details).
As precisely tracking the crop growth on the ground
requires high-spatial-resolution satellite images,
we partition a county into multiple grids 
at the resolution of $9$x$9$ km,
with each grid corresponding to one satellite image.
Figures~\ref{fig:dataset-grid-example} and \ref{fig:dataset-grid-example-ag} 
illustrates an example of county partitioning (refer to Appendix~\ref{sup:county-partition} 
for more details).
The downloaded satellite images for one U.S. state (including all counties therein) spanning one season
are stored in one Hierarchical Data Format (HDF5) file.
Three reasons motivate us to employ the HDF5 file format.
First, it can significantly save the hard disk space.
That is, 
the collected satellite images with a total of $4562.2$ GB  
shrank to $2326.7$ GB (\ie\ $0.51$x smaller space occupancy) in the HDF5 file.
This can facilitate researchers and practitioners 
for lower hard disk space requirements and faster data retrieval. 
Second, it allows for storing data in the form of multidimensional arrays,
making satellite images easy to access.
The HDF5 file for Sentinel-2 Imagery is organized in the form of $(F, T, G, H, W, C)$,
where $F$ represents the FIPS code (\ie\ the unique number for each U.S. county) 
used for retrieving one county's data,
$T$ indicates the number of temporal data in a $14$-day interval with respect to one season,
$G$ represents the number of high-resolution grids for a county,
and $(H, W, C)$ are the width, height, and channel numbers for the satellite image.
Third, it can store descriptive information for the satellite image, such as
its revisit day, the latitude and longitude information it represents, among others.

\begin{table*} [!t] 
    \scriptsize
    \centering
    \setlength\tabcolsep{10 pt}
    \caption{
        Details of WRF-HRRR Computed Dataset
        }
        \begin{tabular}{@{}ccc@{}}
            \toprule
            Source                          & Parameters                  & Description                                                                   \\ \midrule
            \multirow{6}{*}{WRF-HRRR model} & Averaged Temperature        & 2 metre averaged temperature during a day/month. Unit: K                      \\
                                            & Precipitation               & Total precipitation. Unit: $\rm{kg} / \rm{m}^{2}$                                          \\
                                            & Relative Humidity           & 2 metre relative humidity. Unit: \%                                           \\
                                            & Wind Gust                   & Wind gust on the ground. Unit: $\rm{m} / \rm{s}$                                       \\
                                            & Wind Speed                  & Wind speed on the ground. Unit: $\rm{m} / \rm{s}$                                       \\
             &
              Downward Shortwave Radiation Flux &
              \begin{tabular}[c]{@{}c@{}} The total amount of shortwave radiation \\ that reaches the Earth’s surface. Unit: $\rm{W} / \rm{m}^{2}$     \end{tabular} \\ \midrule
            \multirow{3}{*}{Computed by us}           & Maximal Temperature         & 2 metre maximal temperature during a day/month. Unit: K                       \\
                                            & Minimal Temperature         & 2 metre minimal temperature during a day/month. Unit: K                       \\
                                            & Vapor Pressure Deficit (VPD) & The amount of drying power the air has upon the plant. Unit: kPa \\ \bottomrule 
            \end{tabular}
    \label{tab:dataset-wrf-hrrr}
\end{table*}

\par\smallskip\noindent
{\bf WRF-HRRR Computed Dataset.} 
The WRF-HRRR Computed Dataset
is sourced from the WRF-HRRR model~\cite{hrrr},
which produces GRID files on the hourly basis,
containing meteorological parameters over the contiguous U.S. continent
at a spatial resolution of $3$x$3$ km.
To lift the domain knowledge required for using the WRF-HRRR data,
our CropNet dataset includes $9$ carefully chosen and crop growth-relevant meteorological parameters,
with $6$ parameters obtained directly from the WRF-HRRR model,
\ie\ averaged temperature, precipitation, relative humidity, wind gust, wind speed, downward shortwave radiation flux,
and other $3$ parameters computed by ourselves,
\ie\ maximal temperature, minimal temperature, vapor pressure deﬁcit (VPD).
Table~\ref{tab:dataset-wrf-hrrr} presents details of meteorological parameters in the WRF-HRRR Computed Dataset.
%
Notably,
VPD describes the difference between the amount of moisture in the air 
and the maximum amount of moisture the air can hold at a specific temperature,
which is an important concept in understanding 
the environmental conditions that affect plant growth and transpiration.
Given two meteorological parameters,
\ie\ the temperature measured in Kelvin $T_{K}$ 
and the relative humidity $RH$,
VPD is calculated by the following equations:
\begin{equation} \label{eq:vpd}
  \begin{gathered}
    T_{C} = T_{K} - 273.15, \\
    VP_{\rm{sat}} = \frac{610.7 \times 10^{(7.5 \times T_{C}) / (237.3 + T_{C})}}{1000}, \\
    VP_{\rm{air}} = VP_{\rm{sat}} \times \frac{RH}{100}, \\
    VPD = VP_{\rm{sat}} - VP_{\rm{air}}.     
  \end{gathered}
\end{equation}



Two challenges impede us from efficiently and effectively extracting meteorological parameters from GRID files.
First, the resolution in the WRF-HRRR Computed Dataset 
should align with the one in the Sentinel-2 Imagery, \ie\ $9$x$9$ km\footnote{Note that acquiring satellite images at a spatial resolution of $3$x$3$km is infeasible in practice due to its tremendous space size requirement (\ie\ over $20$ TB).}.
A novel solution is proposed to address this issue.
We first follow the Sentinel-2 Imagery by partitioning one county into multiple grids 
at the spatial resolution of $9$x$9$ km.
Then, we utilize the latitude and longitude of the centric point in the $9$x$9$km grid
to find the nearest $3$x$3$km grid in the WRF-HRRR model.
Next, meteorological parameters in the $3$x$3$ km grid 
and its surrounding $8$ grids can be used for representing a region gridded at $9$x$9$ km, 
as shown in Figure~\ref{fig:method-hrrr-grid}. 
In this way, our dataset allows researchers to 
capture the immediate effects of atmospheric weather variations 
occurring directly above the crop-growing area on crop yields.
Second, extracting meteorological parameters from GRID files is extremely time-consuming
as searching the nearest grids requires to 
match geo-grids across the continental United States.
To handle this challenge,
we develop a global cache solution by pre-storing the nearest grid information
corresponding to a pair of latitude and longitude for each location,
reducing the required extraction time from $60$ days to $42$ days
(\ie\ $1.42$x faster than the one without global caching).

The daily meteorological parameters are computed out of the hourly data extracted from the GRID file,
while the monthly weather parameters are derived from our daily data 
to significantly reduce the frequency of accessing the GRID file.
Finally, daily and monthly meteorological parameters are stored in the Comma Separated Values (CSV) file, 
making them readable by researchers and accessible for deep learning models.
The CSV file also includes additional valuable information such as the FIPS code of a county and the latitude and longitude of each grid. 
This provides easy and convenient access to relevant data for researchers.

\par\smallskip\noindent
{\bf USDA Crop Dataset.} 
The data in the USDA Crop Dataset is retrieved from the USDA Quick Statistic website~\cite{usda} via our newly developed web crawler solution.
For each crop type,
the USDA website provides its crop information at the county level in a one-year interval,
with a unique key for identifying the data for one crop type per year, 
\eg\ ``85BEE64A-E605-3509-B60C-5836F6FBB5F6'' for the corn data in 2022.
Our web crawler first retrieves the unique key by specifying the crop type and the year we need.
Then, it utilizes the unique key to obtain the corresponding crop data in one year.
Finally, the downloaded crop data is stored in the CSV file.
Notably, other useful descriptive information, \eg\ FIPS code, state name, county name, \etc,
are also contained in the CSV file for facilitating readability and accessibility.

However, the crop statistic data from the USDA Quick Statistic website is not deep learning-friendly.
For example, 
it uses two columns, \ie\ ``Data Item'' and ``Value'', 
to keep all valuable crop information.
That is, if the description of the ``Data Item'' column refers to the corn yield, 
then the numerical data in the ``Value'' column represents the corn yield.
Otherwise, the data in ``Value'' may signify other information, 
\eg\ the corn production, the soybeans yield, \etc\
New data pre-processing techniques are developed to unify the data format,
making the production and yield information stored in two independent columns
for facilitating Python libraries (\eg\ pandas) to access them.

Our CropNet dataset specifically targets county-level crop yield predictions 
across the contiguous U.S. continent.
We utilize the FIPS code to rapidly fetch the data of each county,
including a list of HDF5 files for Sentinel-2 Imagery,
two lists of CVS files respectively for daily and monthly meteorological parameters,
and one CVS file for the USDA Crop Dataset,
with configurations stored in the JSON file for increasing accessibility. 
(see Appendix~\ref{sup:json-file} for an example of our JSON configuration file).

\section{Experiments and Results}
\label{sec:exp}

Three scenarios of climate change-aware crop yield predictions,
\ie\ \textbf{Crop Yield Predictions}, \textbf{One-Year Ahead Predictions}, 
and \textbf{Self-Supervised Pre-training},
are considered to exhibit the general applicability of our CropNet dataset
to various types of deep learning solutions.
%


\begin{table*} [!t] 
    \scriptsize
    \centering
    \setlength\tabcolsep{6 pt}
    \caption{
        Overall performance for $2022$ crop yield predictions, 
        where the yield of cotton is measured in pounds per acre (LB/AC) 
        and those of the rest are measured in bushels per acre (BU/AC)
        }
        \begin{tabular}{@{}ccccccccccccc@{}}
            \toprule
            \multirow{2}{*}{Method} & \multicolumn{3}{c}{Corn} & \multicolumn{3}{c}{Cotton} & \multicolumn{3}{c}{Soybeans} & \multicolumn{3}{c}{Winter   Wheat} \\ \cmidrule(lr){2-4} \cmidrule(lr){5-7} \cmidrule(lr){8-10} \cmidrule(lr){11-13}
                     & RMSE ($\downarrow$) & R$^{2}$ ($\uparrow$)    & Corr ($\uparrow$)  & RMSE ($\downarrow$) & R$^{2}$ ($\uparrow$)    & Corr ($\uparrow$)  & RMSE ($\downarrow$) & R$^{2}$ ($\uparrow$)    & Corr ($\uparrow$)  & RMSE ($\downarrow$) & R$^{2}$ ($\uparrow$)    & Corr ($\uparrow$)  \\ \midrule
            ConvLSTM & 19.2 & 0.795 & 0.892 & 56.7 & 0.834 & 0.913   & 5.3 & 0.801 & 0.895 & 6.0 & 0.798 & 0.893 \\
            CNN-RNN  & 14.3 & 0.867 & 0.923 & 54.5 & 0.826 & 0.899   & 4.1 & 0.853 & 0.915 & 5.6 & 0.823 & 0.906 \\
            GNN-RNN  & 14.1 & 0.871 & 0.917 & 55.1 & 0.813 & 0.881   & 4.1 & 0.868 & 0.929 & 5.3 & 0.845 & 0.912 \\
            MMST-ViT & 13.2 & 0.890  & 0.943 & 50.9 & 0.848 & 0.921  & 3.9 & 0.879 & 0.937 & 4.8 & 0.864 & 0.929 \\ \bottomrule
        \end{tabular}
        \label{tab:exp-overall}
\end{table*}

\subsection{Experimental Settings}
\label{sec:exp-setup}

\par\smallskip\noindent
{\bf Approaches.} 
The LSTM-based, CNN-based, GNN-based, and ViT-based models
are represented respectively by \textbf{ConvLSTM}~\cite{shi:nips15:conv_lstm},
\textbf{CNN-RNN}~\cite{khaki2020cnn}, \textbf{GNN-RNN}~\cite{fan:aaai23:crop_prediction},
and \textbf{MMST-ViT}~\cite{fudong:iccv23:mmst_vit} in our experiments,
targeting crop yield predictions.
Meanwhile, two self-supervised learning (SSL) techniques,
\ie\ \textbf{MAE}~\cite{he:cvpr22:mae}, 
and \textbf{MM-SSL} in the MMST-ViT,
serving respectively as unimodal and multi-modal SSL techniques,
are taken into account under the self-supervised pre-training scenario.
The aforementioned methods are modified slightly to make them fit the CropNet data in our experiments.
%

\par\smallskip\noindent
{\bf Metrics.} 
Three performance metrics, \textit{i.e.}, \textbf{Root Mean Square Error (RMSE)}, 
\textbf{R-squared (R$^{2})$}, 
and \textbf{Pearson Correlation Coefficient (Corr)},
are adopted to evaluate the efficacy of the CropNet dataset for crop yield predictions. 
Note that a lower RMSE value and a higher R$^{2}$ (or Corr) value represent better prediction performance.

Details of utilizing our CropNet data for conducting experiments 
are deferred to Appendix~\ref{sup:exp-setup}
for conserving space.

\subsection{Performance Evaluation for 2022 Crop Yield Predictions}
\label{sec:exp-overall}

We conduct experiments on the CropNet dataset for 2022 crop yield predictions 
by using satellite images and daily weather conditions during growing seasons,
as well as monthly meteorological conditions from 2017 to 2021,
running under the  ConvLSTM, CNN-RNN, GNN-RNN, and MMST-ViT models. 
Table~\ref{tab:exp-overall} presents each crop's  overall performance results 
(\ie\ RMSE, R$^{2}$, and Corr) 
in aggregation.
%
We have two observations.
First, all models achieve superb prediction performance with our CropNet data.
For example, ConvLSTM, CNN-RNN, GNN-RNN, and MMST-ViT achieve small RMSE values
of $5.3$, $4.1$, $4.1$, and $3.9$, respectively, for soybeans yield predictions (see the 8th column).
These results validate that our CropNet dataset is well-suited 
for LSTM-based, CNN-based, and GNN-based, and ViT-based models,
demonstrating its general applicability. 
Second, MMST-ViT achieves the best performance results under all scenarios,
with lowest RMSE values of $13.2$, $50.9$, $3.9$, and $4.8$,
as well as highest R$^{2}$ (or Corr) values of 
$0.890$ (or $0.943$), $0.848$ (or $0.921$), $0.879$ (or $0.937$), and $0.864$ (or $0.929$),
respectively for predicting corn, cotton, soybeans, and winter wheat yields. 
This is due to MMST-ViT's novel attention mechanisms~\cite{lyu:amia22:attn,vaswani:nips17:attention,lyu:naacl22:attn,wang2024medformer,he:nips23:atten}, 
which perform the cross-attention 
between satellite images and meteorological parameters, 
able to capture the effects of both growing season weather variations
and climate change on crop growth.
This experiment exhibits that our CropNet dataset can provide crop yield predictions timely and precisely, 
essential for making informed economic decisions, optimizing agricultural resource allocation, \etc\

\begin{figure*} [!t] 
    \captionsetup[subfigure]{justification=centering}
    \begin{subfigure}[t]{0.32\textwidth}
        \centering
        \includegraphics[width=\textwidth]{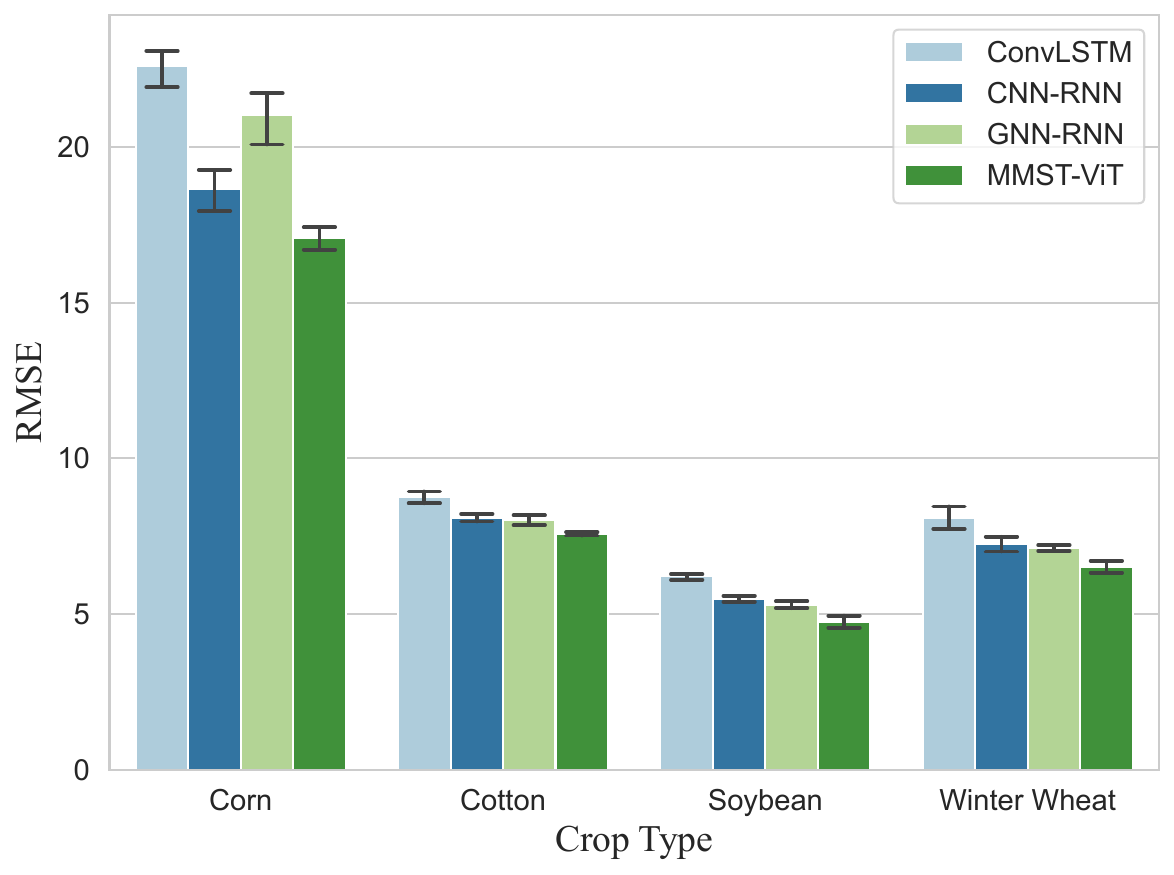}
        \caption{RMSE}
        \label{fig:exp-early-pred-rmse}
    \end{subfigure}
    \begin{subfigure}[t]{0.32\textwidth}
        \centering
        \includegraphics[width=\textwidth]{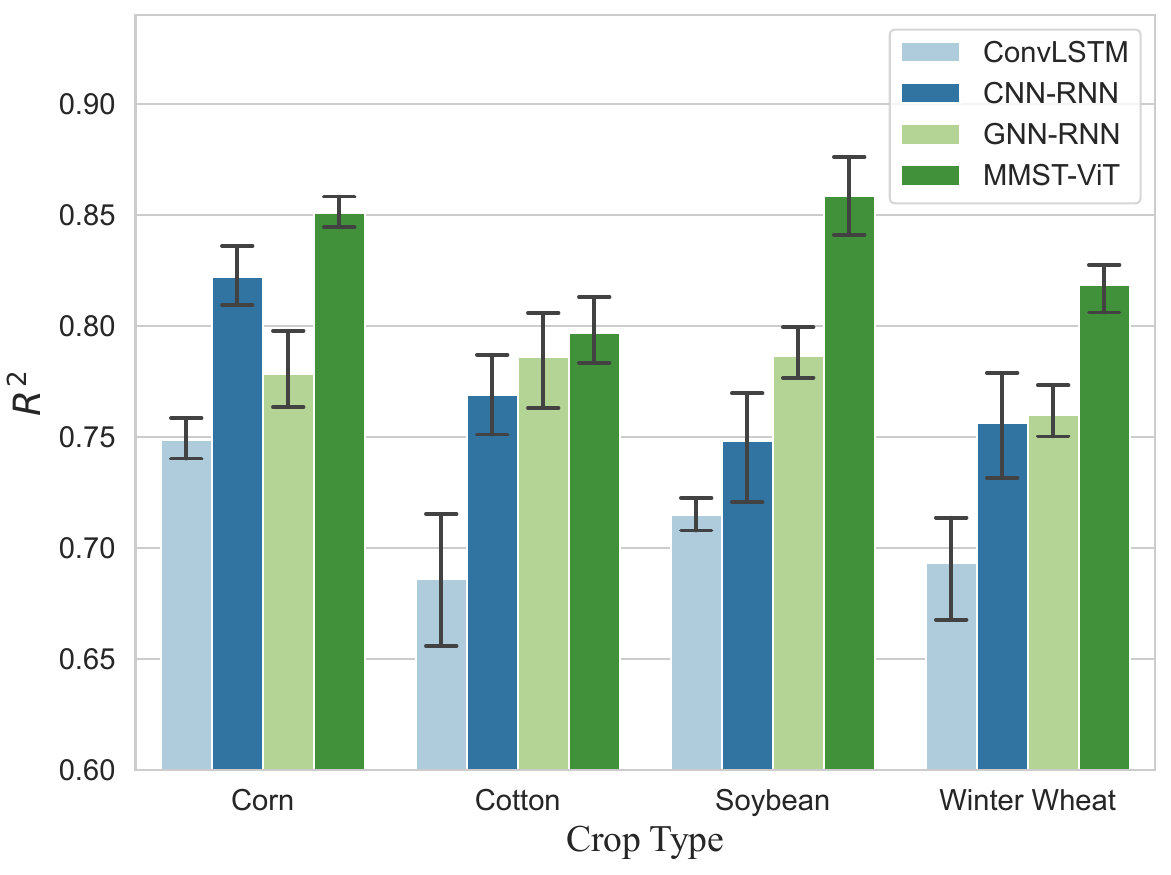}
        \caption{R$^2$}
        \label{fig:exp-early-pred-r2}
    \end{subfigure}
    \begin{subfigure}[t]{0.32\textwidth}
        \centering
        \includegraphics[width=\textwidth]{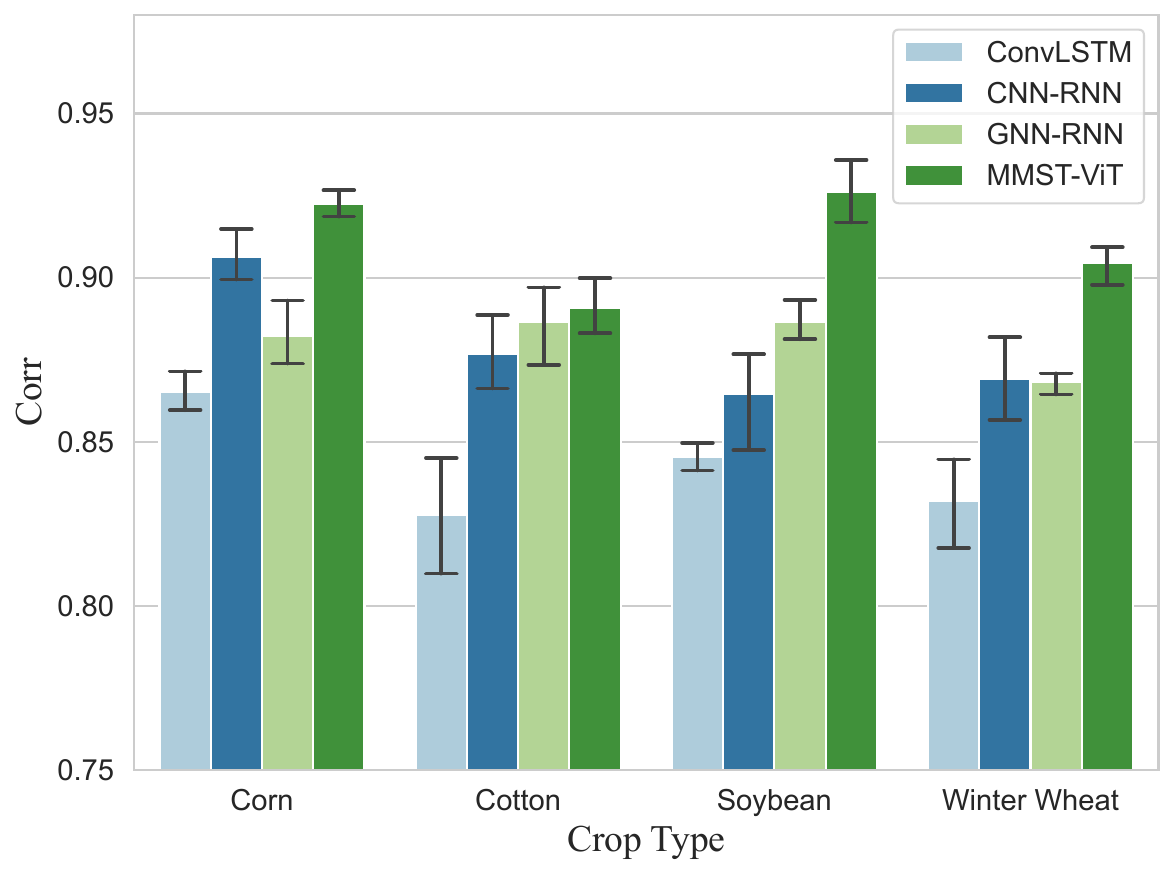}
        \caption{Corr}
        \label{fig:exp-early-pred-corr}
    \end{subfigure}
    \vspace{-0.8 em}
    \caption{
        The performance of one-year ahead crop yield predictions, 
        with the cotton yield measured by LB/AC and other crop yields measured by BU/AC.
        In Figure~\ref{fig:exp-early-pred-rmse}, 
        we present the square root of the RMSE values for the cotton yield 
        to enhance visualization.
    }
    \label{fig:exp-early-pred}
    \vspace{-0.5 em}
\end{figure*}

\subsection{Performance of One-Year Ahead Predictions}
\label{sec:exp-one-year}

Crop yield predictions well in advance of the planting season 
are also critical for farmers to make early crop planting and management plans. 
Here, we apply the CropNet dataset one year before the planting season for predicting the next year's crop yields.
Figure~\ref{fig:exp-early-pred} shows our experimental results for $2022$ crop yield predictions 
by using our CropNet data during the $2021$ growing season.
We observe that all models can still maintain decent prediction performance.
For instance, ConvLSTM, CNN-RNN, GNN-RNN, and MMST-ViT achieve the averaged RMSE values 
of $6.2$, of $5.4$, of $5.3$, and of $4.7$, respectively, for soybeans predictions.
Meanwhile, MMST-ViT consistently achieves excellent Corr values, 
averaging at $0.922$ for corn, $0.890$ for cotton, $0.926$ for soybeans, and $0.904$ for winter wheat predictions,
only slightly inferior to the performance results for the regular 2022 crop yield predictions 
(see the last row in Table~\ref{tab:exp-overall}).
This can be attributed to MMST-ViT's ability to capture the indirect influence of 2021's weather conditions 
on crop growth in the subsequent year through the utilization of long-term weather parameters, 
which further underscores how our CropNet dataset enhances climate change-aware crop yield predictions.
%

\subsection{Improving the Generalization Capabilities of DNNs}
\label{sec:exp-ssl}

\begin{figure*} [!t] 
    \captionsetup[subfigure]{justification=centering}
    \begin{subfigure}[t]{0.32\textwidth}
        \centering
        \includegraphics[width=\textwidth]{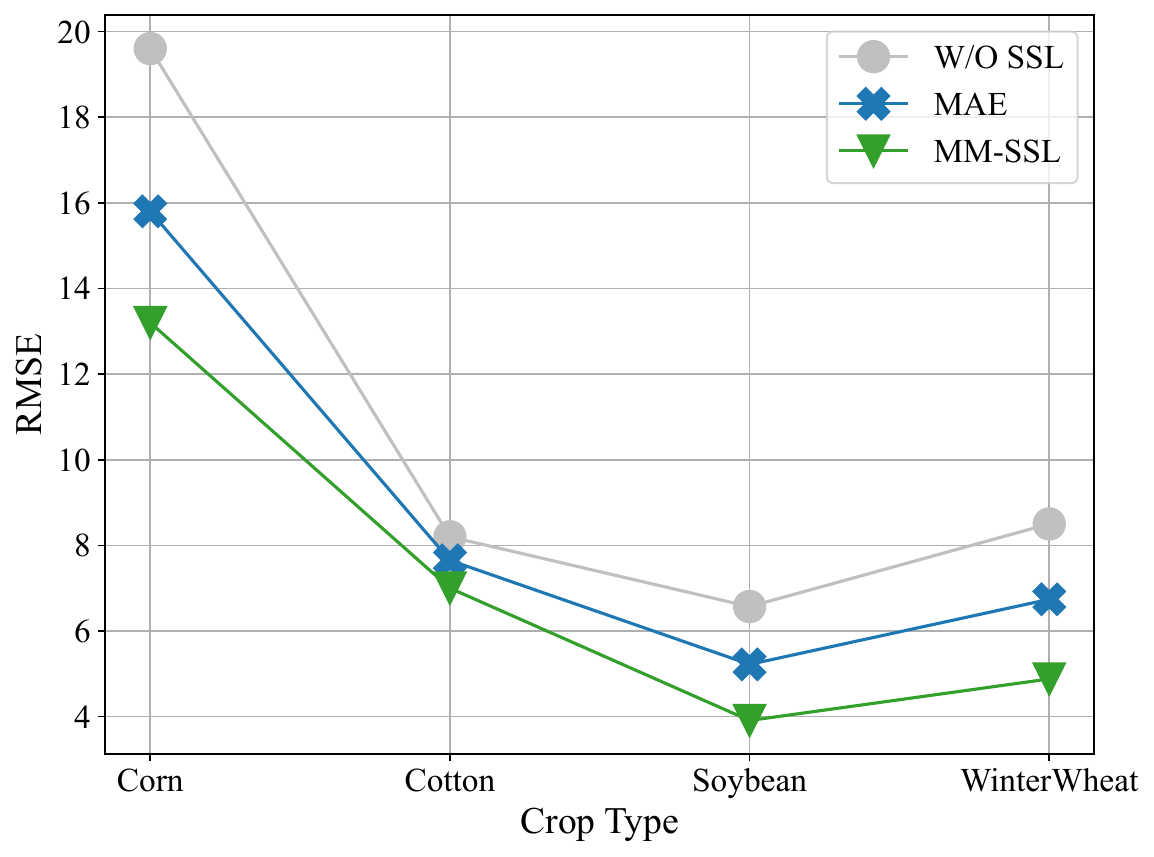}
        \caption{RMSE}
        \label{fig:exp-ssl-rmse}
    \end{subfigure}
    \begin{subfigure}[t]{0.32\textwidth}
        \centering
        \includegraphics[width=\textwidth]{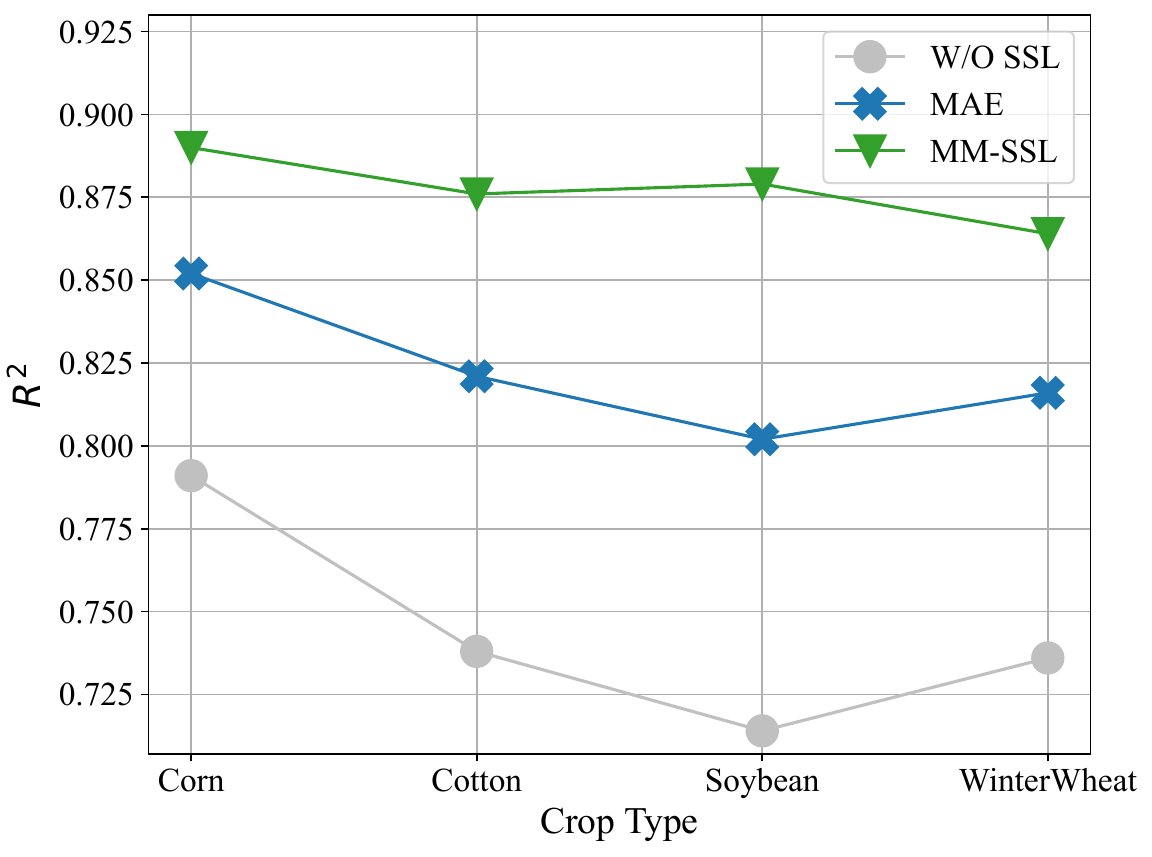}
        \caption{R$^2$}
        \label{fig:exp-ssl-r2}
    \end{subfigure}
    \begin{subfigure}[t]{0.32\textwidth}
        \centering
        \includegraphics[width=\textwidth]{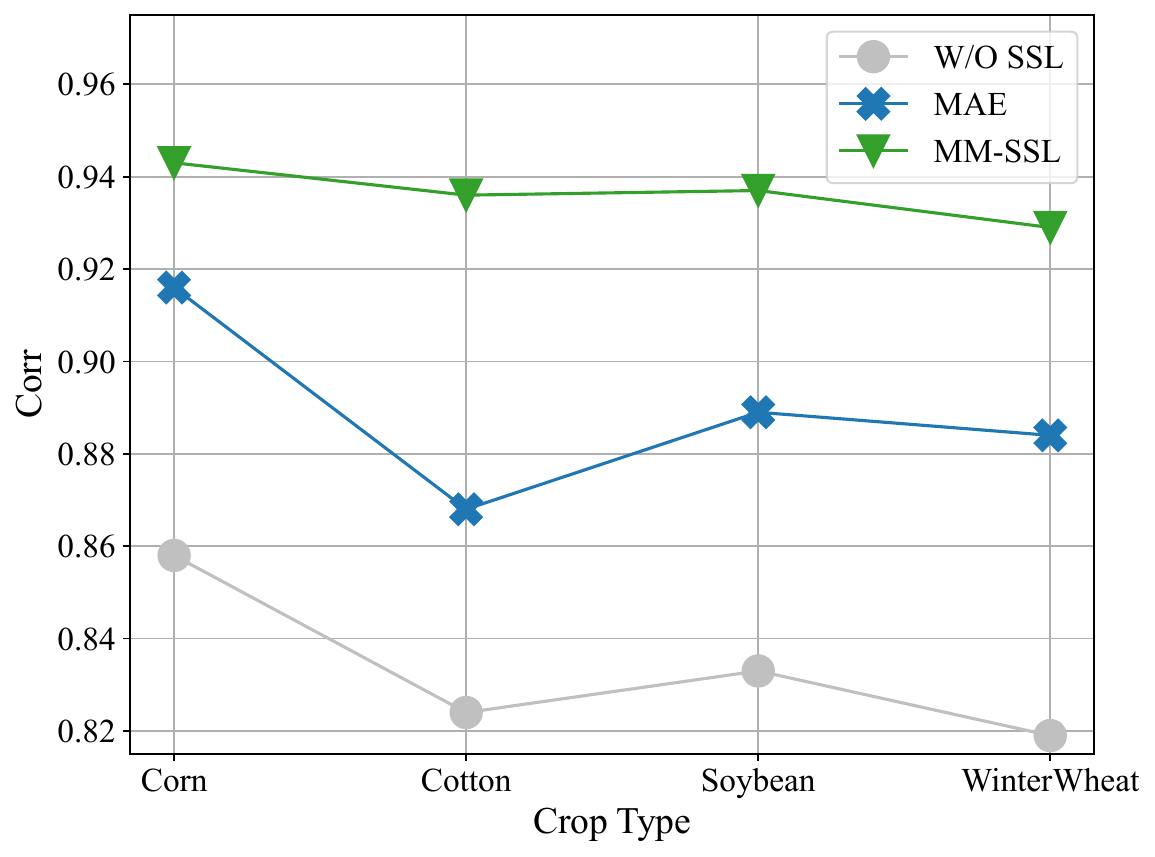}
        \caption{Corr}
        \label{fig:exp-ssl-corr}
    \end{subfigure}
    \vspace{-0.5 em}
    \caption{
        Illustration of how our CropNet dataset benefits self-supervised learning techniques.
        Notably, Figure~\ref{fig:exp-ssl-rmse} depicts the square root of RMSE values for the cotton yield to improve visualization.
    }
    \label{fig:exp-ssl}
\end{figure*}

Self-supervised learning (SSL) techniques~\cite{chen:icml20:simclr,bao:iclr22:beit,he:cvpr22:mae,zhou:ijcnn23:ssl,zhang:cvprw23:mm_bsn,wang2023metamix,wang:nips23:ssl} 
have significantly advanced the generalization capabilities of deep neural networks (DNNs), 
especially in vision transformers (ViTs).
Our CropNet dataset with a total size of over $2$ TB of data
can benefit both deep-learning and agricultural communities
by providing large-scale visual satellite imagery and numerical meteorological data 
for pre-training DNNs.
To exhibit the applications of our CropNet dataset to self-supervised pre-training,
we adopt the MMST-ViT for crop yield predictions by considering three scenarios,
\ie\ MMST-ViT without the SSL technique (denoted as ``w/o SSL''), 
MMST-ViT with the SSL technique in MAE (denoted as ``MAE''),
and MMST-ViT with the multi-modal SSL technique 
proposed in \cite{fudong:iccv23:mmst_vit} (denoted as ``MM-SSL'').
Figure~\ref{fig:exp-ssl} illustrates the performance results for four crop types
under three performance metrics of interest (\ie\ RMSE, R$^{2}$, and Corr).
We discover that without the SSL technique (\ie\ the gray line), 
the MMST-ViT model exhibits limitations in generalization capabilities,
resulting in suboptimal crop yield prediction performance across all tested scenarios.
Besides, pre-training MMST-ViT with the SSL technique in MAE (\ie\ the blue line) 
improves its performance results (compared to the ``w/o SSL''),
with decreased RMSE values by $3.8$, $9.6$, $1.3$, and $1.7$ 
for corn, cotton, soybeans, and winter wheat predictions, respectively.
This statistical evidence confirms that our CropNet dataset 
can improve the generalization capabilities in vision models.
Furthermore, MMST-ViT with the multi-modal SSL technique (\ie\ the green line)
achieves the best performance results under all scenarios.
In comparison to the ``w/o SSL'' scenario, it decreases RMSE values by $6.4$, $18.3$, $2.6$, and $3.6$,
respectively, for predicting corn, cotton, soybeans, and winter wheat.
The effectiveness of the multi-modal SSL technique 
may stem from its ability to integrate visual satellite imagery with numerical meteorological data 
found in the CropNet dataset.
This integration enhances the generalization capabilities of the MMST-ViT model
by improving its ability to effectively discern the influence of weather conditions 
on crop growth patterns during the pre-training phase.


\begin{table*} [!t] 
    \scriptsize
    \centering
    \setlength\tabcolsep{12 pt}
    \caption{
        Ablation studies for different modalities of the CropNet dataset, 
        with five scenarios considered and the last row 
        presenting the results by using all modalities
        }
        \vspace{-0.5em}
        \begin{tabular}{@{}cccccccc@{}}
            \toprule
            \multirow{2}{*}{Modality}           & \multirow{2}{*}{Scenario} & \multicolumn{3}{c}{Corn} & \multicolumn{3}{c}{Soybeans} \\ \cmidrule(l){3-5} \cmidrule(l){6-8} 
                                                &                           & RMSE ($\downarrow$)   & R$^{2}$ ($\uparrow$)     & Corr ($\uparrow$)  & RMSE ($\downarrow$)    & R$^{2}$ ($\uparrow$)      & Corr ($\uparrow$)    \\ \midrule
            \multirow{2}{*}{Sentinel-2 Imagery} & w/o temporal images       & 22.1   & 0.758  & 0.870  & 5.72    & 0.773    & 0.879   \\
                                                & w/o high-resolution images       & 27.9   & 0.656  & 0.810  & 7.80    & 0.631    & 0.794   \\ \midrule
            \multirow{3}{*}{\begin{tabular}[c]{@{}c@{}}WRF-HRRR \\ Computed Dataset\end{tabular}} & w/o WRF-HRRR data & 20.6 & 0.758 & 0.871 & 5.78 & 0.764 & 0.874 \\
                                                & w/o short-term data       & 18.6   & 0.796  & 0.892  & 5.04    & 0.816    & 0.903   \\
                                                & w/o long-term data        & 15.3   & 0.854  & 0.924  & 4.72    & 0.825    & 0.908   \\ \midrule
            All                                 & —                          & 13.2   & 0.890   & 0.943  & 3.91    & 0.879    & 0.937   \\ \bottomrule
            \end{tabular}
        \label{tab:exp-as}
    \vspace{-0.5 em}
\end{table*}

\subsection{Significance of Each Modality of Our CropNet Dataset}
\label{sec:exp-as}

To show the necessity and significance of each modality data in our CropNet dataset, 
we examine five scenarios. 
First, we drop the temporal satellite images (denoted as ``w/o temporal images'')
by randomly selecting only one day's imagery data.
Second, we discard the high-resolution satellite image (denoted as ``w/o high-resolution images'')
by using only one satellite image to capture the whole county's agricultural information.
Third, we ignore the effects of weather variations on crop yields 
by dropping all meteorological data, denoted as ``w/o WRF-HRRR data''.
Similarly, ``w/o short-term data'' and ``w/o long-term data'' represent 
masking out the daily and monthly meteorological parameters, respectively.
We also include prediction results
by using all modalities of the CropNet (denoted as ``All'') for performance comparison.
Note that the USDA Crop Dataset provides the label for crop yield predictions; 
hence, no ablation study requires.

Table~\ref{tab:exp-as} presents the experimental results 
under the MMST-ViT model~\cite{fudong:iccv23:mmst_vit}. 
We have four observations.
First, discarding the temporal satellite images (\ie ``w/o temporal images'') 
degrades performance significantly, raising the RMSE value by $8.9$ (or $1.81$) 
and lowering the Corr value by $0.073$ (or $0.058$) 
for corn (or soybeans) yield predictions.
This is due to that a sequence of satellite images spanning the whole growing season 
are essential for tracking crop growth.
Second, ``w/o high-resolution images'' achieves the worst prediction performance,
with a largest RMSE vaue of $27.9$ (or $7.8$) 
and a lowest Corr value of $0.810$ (or $0.794$) for corn (or soybeans) yield predictions.
The reason is that high-resolution satellite images are critical for precise agricultural tracking.
Third, dropping meteorological parameters (\ie\ w/o WRF-HRRR data) 
makes MMST-ViT fail to capture meteorological effects on crop yields,
leading to the increase of RMSE value by $7.4$ (or $1.87$)
and the decease of Corr value by $0.072$ (or $0.063$) for predicting corn (or soybeans) yields.  
Fourth, discarding either daily weather parameters (\ie ``w/o short-term data'')
or monthly meteorological parameters (\ie ``w/o long-term data'') 
lowers crop yield prediction performance.
The reason is that the former is necessary for capturing growing season weather variations,
while the latter is essential for monitoring long-term climate change effects.
Hence, we conclude that each modality in our CropNet dataset is important and necessary for accurate crop yield predictions,
especially for those crops which are sensitive to growing season weather variations and climate change.


\section{The CropNet Package}
\label{sec:package}

In addition to our CropNet dataset, 
we also release the \textit{CropNet} package,
including three types of APIs,
at the Python Package Index (PyPI),
which is designed to 
facilitate researchers in developing DNNs 
for multi-modal climate change-aware crop yield predictions, with its details presented as follows.
%


\begin{figure} [!thp]
    \centering
    \begin{tabular}{c}
\begin{lstlisting}[language=Python, label={list:data-downloader}] 
# Download the 2023 Sentinel-2 Imagery
downloader.download_Sentinel2(fips_codes=["01003"], years=["2023"])

# Download the 2023 WRF-HRRR Computed data
downloader.download_HRRR(fips_codes=["01003"], years=["2023"])

# Download the 2023 USDA Soybean data
downloader.download_USDA("Soybean", fips_codes=["01003"], years=["2023"])
\end{lstlisting}
\end{tabular}
\vspace{-0.5 em}
\caption{Example of our DataDownloader API.}
\label{list:data-downloader}
\vspace{-0.5 em}
\end{figure}


\begin{figure} [!thp]
    \centering
    \begin{tabular}{c}
\begin{lstlisting}[language=Python, label={list:data-retriever}]
# Retrieve the Sentinel-2 Imagery data for two counties
retriever.retrieve_Sentinel2(fips_codes=["01003","01005"],years=["2022"])
   
# Retrieve the WRF-HRRR Computed data for two counties
retriever.retrieve_HRRR(fips_codes=["01003", "01005"], years=["2022"])

# Retrieve the USDA data for two counties
retriever.retrieve_USDA(fips_codes=["01003", "01005"], years=["2022"])
\end{lstlisting}
\end{tabular}
\vspace{-0.5 em}
\caption{Example of our DataRetriever API.}
\label{list:data-retriever}
\vspace{-1 em}
\end{figure}


\begin{figure} [!thp]
    \centering
    \begin{tabular}{c}
\begin{lstlisting}[language=Python, label={list:data-loader}] 
from torch.utils.data import DataLoader

# The base directory for the CropNet dataset
base_dir = "/mnt/data/CropNet"
# The JSON configuration file
config_file = "data/soybeans_train.json"

# The PyTorch dataloaders for each modality of data
sentinel2_loader = DataLoader(Sentinel2Imagery(base_dir, config_file))
hrrr_loader = DataLoader(HRRRComputedDataset(base_dir, config_file))
usda_loader = DataLoader(USDACropDataset(base_dir, config_file))
\end{lstlisting}
\end{tabular}
\vspace{-0.5 em}
\caption{The PyTorch example of our DataLoader API.}
\label{list:data-loader}
\vspace{-1 em}
\end{figure}

\par\smallskip\noindent
\textbf{DataDownloader.} 
This API allows researchers to download the CropNet data over the time/region of interest on the fly.
For example, given the time and region (\eg\ the FIPS code for one U.S. county) of interest,
Figure~\ref{list:data-downloader} presents how to utilize the DataDownloader API to download the up-to-date CropNet data.
%

\par\smallskip\noindent
\textbf{DataRetriever.} 
This API enables researchers to conveniently obtain the CropNet data 
stored in the local machine (\eg\ after you have downloaded our curated CropNet dataset) 
over the time/region of interest,
with the requested data presented in a user-friendly format.
For instance, Figure~\ref{list:data-retriever} shows 
how to employ the DataRetriever API to obtain the CropNet data for two U.S. counties.

\par\smallskip\noindent
\textbf{DataLoader.} 
This API is designed to assist researchers in their development of DNNs for crop yield predictions. 
It allows researchers to flexibly and seamlessly merge multiple modalities of CropNet data, 
and then expose them through a DataLoader object after performing necessary data preprocessing techniques.
A PyTorch example of using our DataLoader API for training (or testing) DNNs is shown in Figure~\ref{list:data-loader}.

\section{Conclusion}\label{sec:conslusion}
This work presented our crafted CropNet dataset, an open, large-scale, and multi-modal
dataset targeting specifically at county-level crop yield predictions
across the contiguous United States continent.
Our CropNet dataset is composed of three modalities of data,
\ie\ Sentinel-2 Imagery, WRF-HRRR Computed Dataset, and USDA Crop Dataset, containing
high-resolution satellite images, daily and monthly meteorological conditions,
and crop yield information, aligned in both the spatial and the temporal domains. 
Such a dataset is ready for wide use in deep learning, agriculture, and meteorology areas, for developing new solutions and models for crop yield predictions,  with the consideration of both the effects of growing season weather variations and climate change on crop growth. 
Extensive experimental results validate the general applicability of our CropNet dataset
to various types of deep learning models
for both the timely and one-year ahead crop yield predictions.
Besides, the applications of our CropNet dataset to self-supervised pre-training scenarios
demonstrate the dataset's versatile utility in 
improving the generalization capabilities of deep neural networks (DNNs).
In addition to our crafted dataset, 
we have also developed the CropNet package,
which allows researchers and practitioners to (1) construct the CropNet data 
on the fly over the time/region of interest 
and (2) flexibly build their deep learning models for climate change-aware crop yield predictions.
Although our initial goal of crafting the CropNet dataset and developing the CropNet package 
is for precise crop yield prediction, 
we believe its future applicability is broad and deserved further exploration. 
It can benefit the deep learning, agriculture, and meteorology communities, 
in the pursuit of more interesting, critical, and pertinent applications.

\section*{Acknowledgments}
This work was supported in part by NSF under Grants 2019511, 2348452, and 2315613. Any opinions and findings expressed in the paper are those of the authors and do not necessarily reflect the view of funding agencies.

\bibliographystyle{ACM-Reference-Format}
\balance
\bibliography{main}

\appendix

\section*{Outline}
This document provided supplementary materials to support our main paper. 
%
%
Section~\ref{sup:dataset-data-collection} provides details of data collection.
%
Section~\ref{sup:exp-setup} presents additional experimental settings. 
\section{Details of Data Collection}
\label{sup:dataset-data-collection}

\subsection{Significance of Our Cloud Coverage Setting and Revisit Frequency for Sentinel-2 Imagery}
\label{sup:sentinel-2-setting}

\begin{figure*}  [!t] 
  \centering
  \captionsetup[subfigure]{justification=centering}
  \begin{subfigure}[t]{0.13\textwidth}
      \centering
      \includegraphics[width=\textwidth]{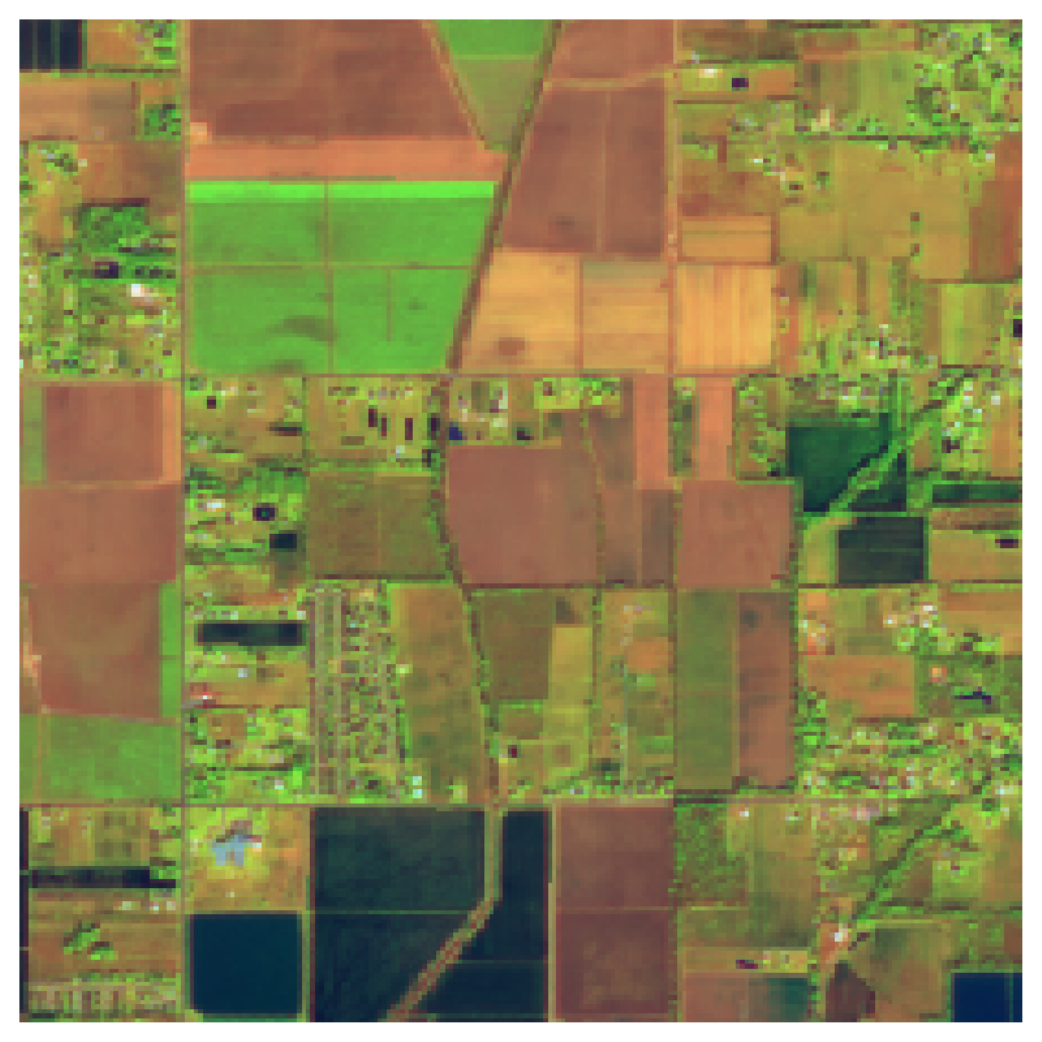}
      \caption{2022-12-01 \\ Cloud: 0\%}
      \label{fig:sup-w-cloud-1201}
  \end{subfigure}
  \begin{subfigure}[t]{0.13\textwidth}
      \centering
      \includegraphics[width=\textwidth]{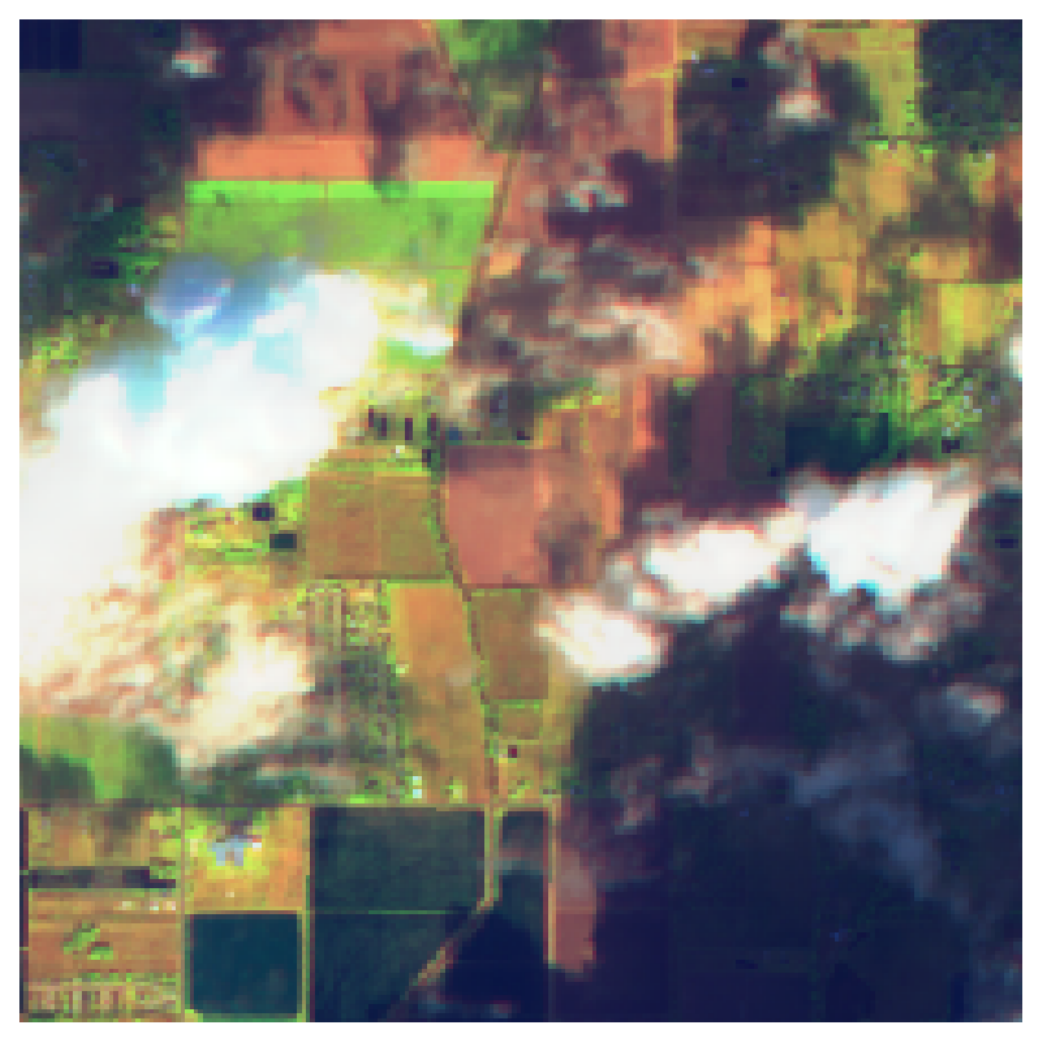}
      \caption{2022-12-06 \\ Cloud: 35.8\%}
      \label{fig:sup-w-cloud-1206}
  \end{subfigure}
  \begin{subfigure}[t]{0.13\textwidth}
    \centering
    \includegraphics[width=\textwidth]{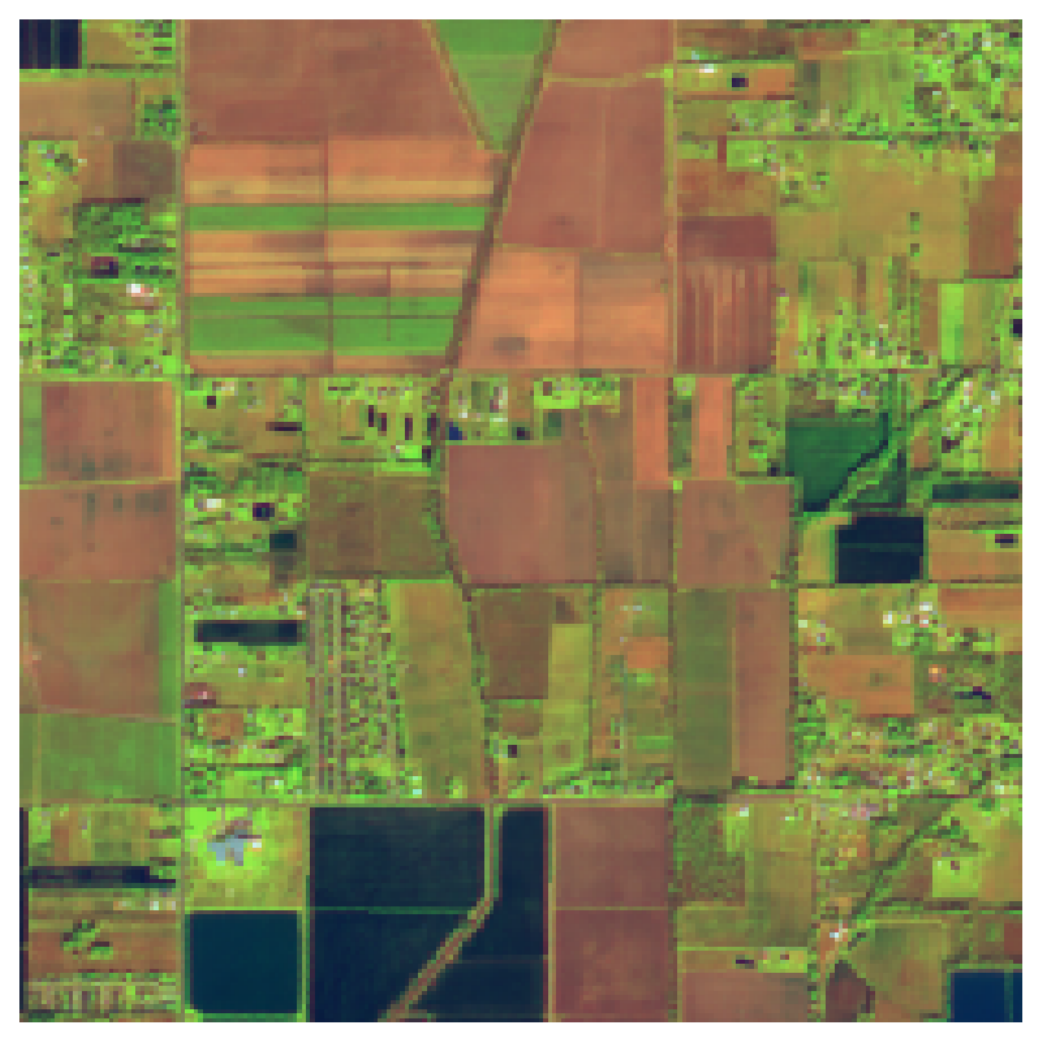}
    \caption{2022-12-11 \\ Cloud: 0\%}
    \label{fig:sup-w-cloud-1211}
  \end{subfigure}
  \begin{subfigure}[t]{0.13\textwidth}
    \centering
    \includegraphics[width=\textwidth]{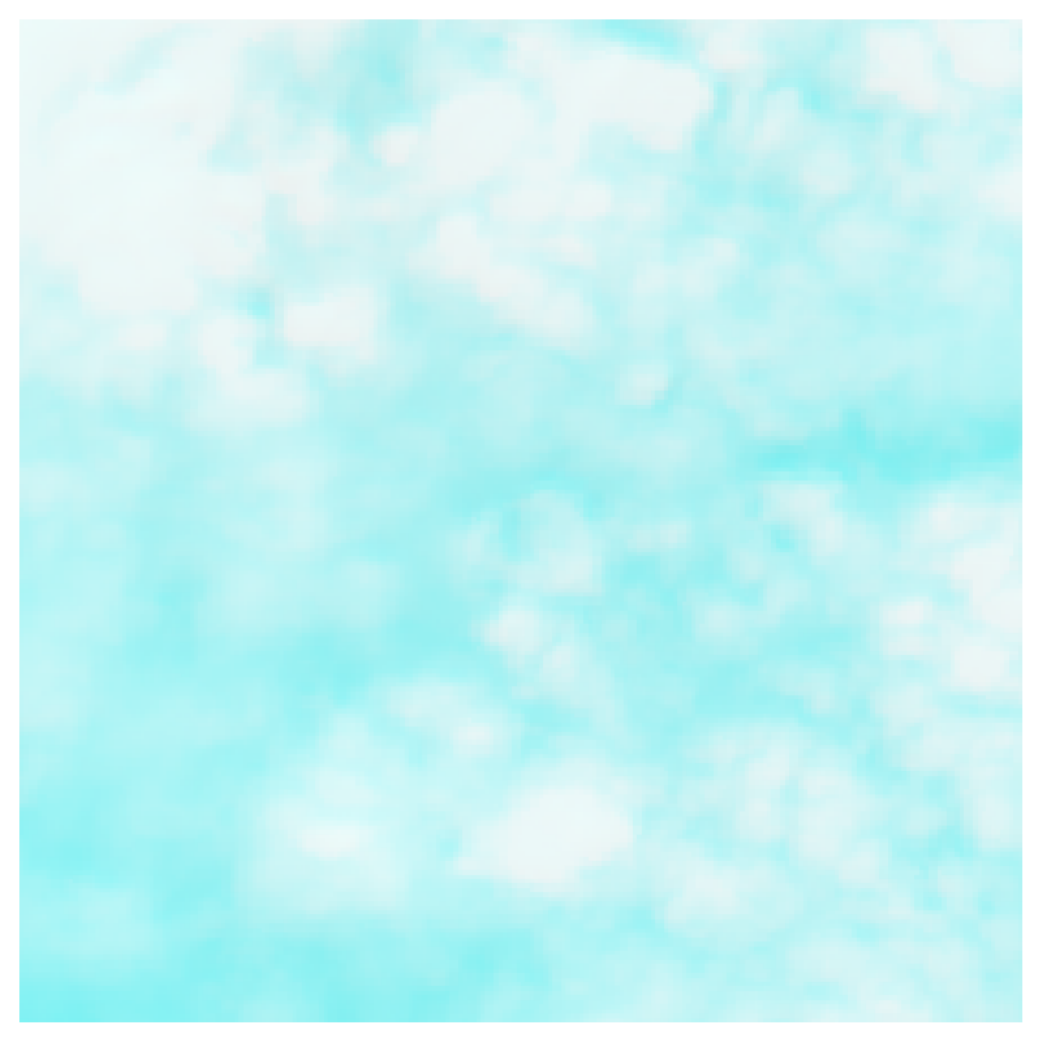}
    \caption{2022-12-16 \\ Cloud: 97.2\%}
    \label{fig:sup-w-cloud-1216}
  \end{subfigure}
  \begin{subfigure}[t]{0.13\textwidth}
    \centering
    \includegraphics[width=\textwidth]{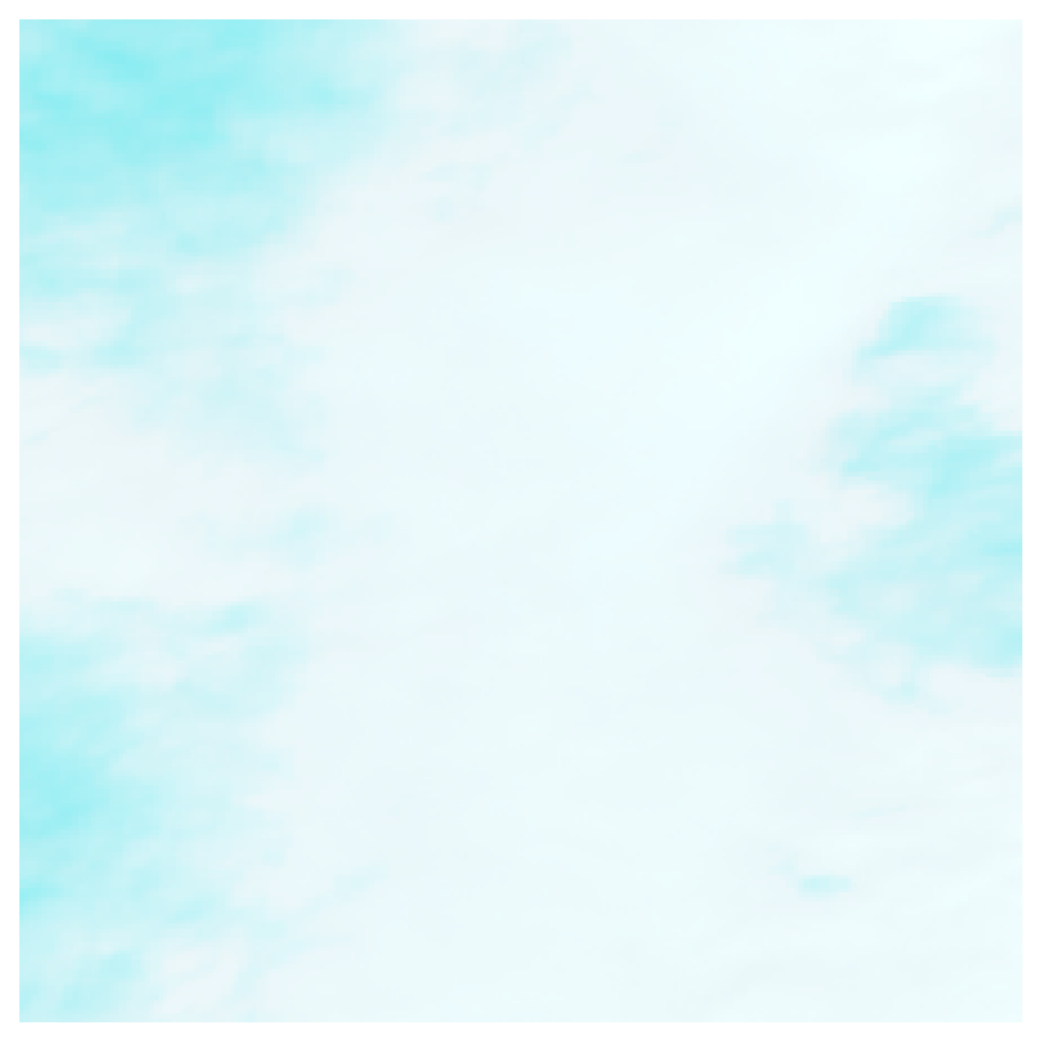}
    \caption{2022-12-21 \\ Cloud: 100\%}
    \label{fig:sup-w-cloud-1221}
  \end{subfigure}
  \begin{subfigure}[t]{0.13\textwidth}
    \centering
    \includegraphics[width=\textwidth]{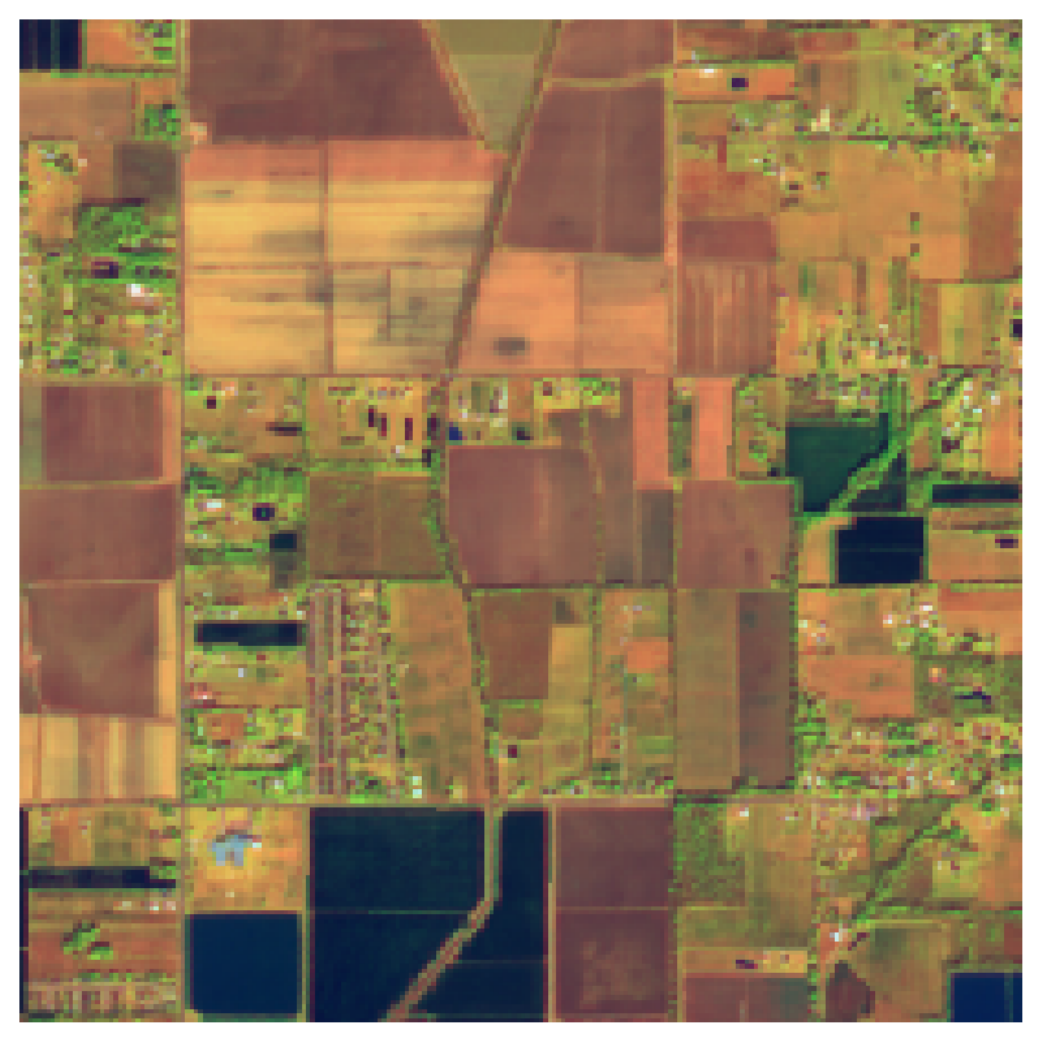}
    \caption{2022-12-26 \\ Cloud: 2.7\%}
    \label{fig:sup-w-cloud-1226}
  \end{subfigure}
  \vspace{-0.5em}
  \caption{
    Examples of Sentinel-2 Imagery under 
    the original revisit frequency of $5$ days 
    without our cloud coverage setting,
    with the revisit date and the cloud coverage listed below each image.
  }
  \label{fig:sup-w-cloud}
  \vspace{-0.5 em}
\end{figure*}

\begin{figure*} [!t]
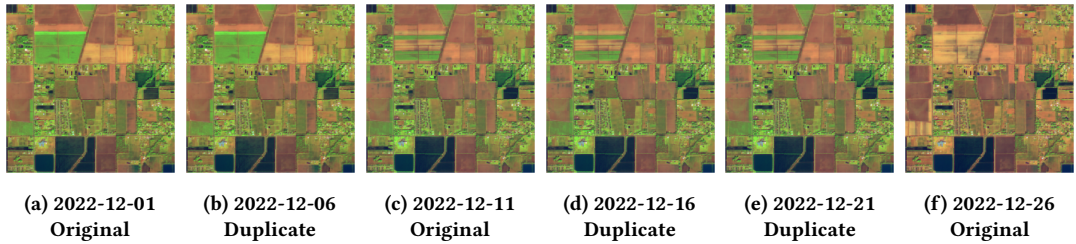
 
  \centering
  \captionsetup[subfigure]{justification=centering}
  \begin{subfigure}[t]{0.13\textwidth}
      \centering
      \includegraphics[width=\textwidth]{figs/sup-wo-cloud-1201.pdf}
      \caption{2022-12-01 \\ Original}
      \label{fig:sup-wo-cloud-1201}
  \end{subfigure}
  \begin{subfigure}[t]{0.13\textwidth}
      \centering
      \includegraphics[width=\textwidth]{figs/sup-wo-cloud-1206.pdf}
      \caption{2022-12-06 \\ Duplicate}
      \label{fig:sup-wo-cloud-1206}
  \end{subfigure}
  \begin{subfigure}[t]{0.13\textwidth}
    \centering
    \includegraphics[width=\textwidth]{figs/sup-wo-cloud-1211.pdf}
    \caption{2022-12-11 \\ Original}
    \label{fig:sup-wo-cloud-1211}
  \end{subfigure}
  \begin{subfigure}[t]{0.13\textwidth}
    \centering
    \includegraphics[width=\textwidth]{figs/sup-wo-cloud-1216.pdf}
    \caption{2022-12-16 \\ Duplicate}
    \label{fig:sup-wo-cloud-1216}
  \end{subfigure}
  \begin{subfigure}[t]{0.13\textwidth}
    \centering
    \includegraphics[width=\textwidth]{figs/sup-wo-cloud-1221.pdf}
    \caption{2022-12-21 \\ Duplicate}
    \label{fig:sup-wo-cloud-1221}
  \end{subfigure}
  \begin{subfigure}[t]{0.13\textwidth}
    \centering
    \includegraphics[width=\textwidth]{figs/sup-wo-cloud-1226.pdf}
    \caption{2022-12-26 \\ Original}
    \label{fig:sup-wo-cloud-1226}
  \end{subfigure}
  \vspace{-0.5em}
  \caption{
    Examples of Sentinel-2 Imagery under 
    the original revisit frequency of $5$ days 
    and our cloud coverage setting.
    The revisit date is listed below each image.
    ``Duplicate'' (or ``Original'') indicates whether the satellite image 
    is duplicate (or not) under our cloud coverage setting.
  }
  \label{fig:sup-wo-cloud}
  \vspace{-0.5 em}
\end{figure*}

\begin{figure*}  [!t]  
  \centering
  \captionsetup[subfigure]{justification=centering}
  \begin{subfigure}[t]{0.13\textwidth}
      \centering
      \includegraphics[width=\textwidth]{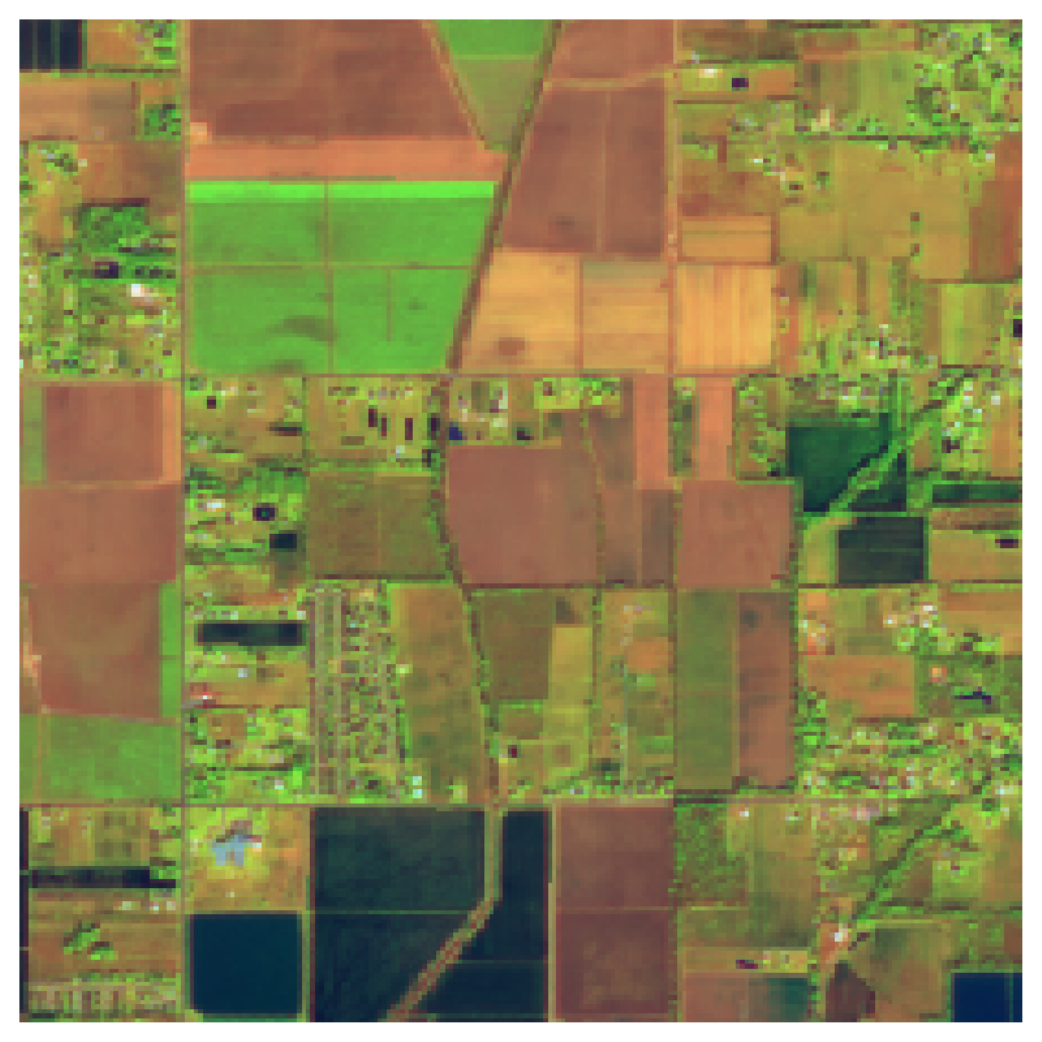}
      \caption{2022-12-01 \\ Original}
      \label{fig:sup-revist-ours-1201}
  \end{subfigure}
  \begin{subfigure}[t]{0.13\textwidth}
      \centering
      \includegraphics[width=\textwidth]{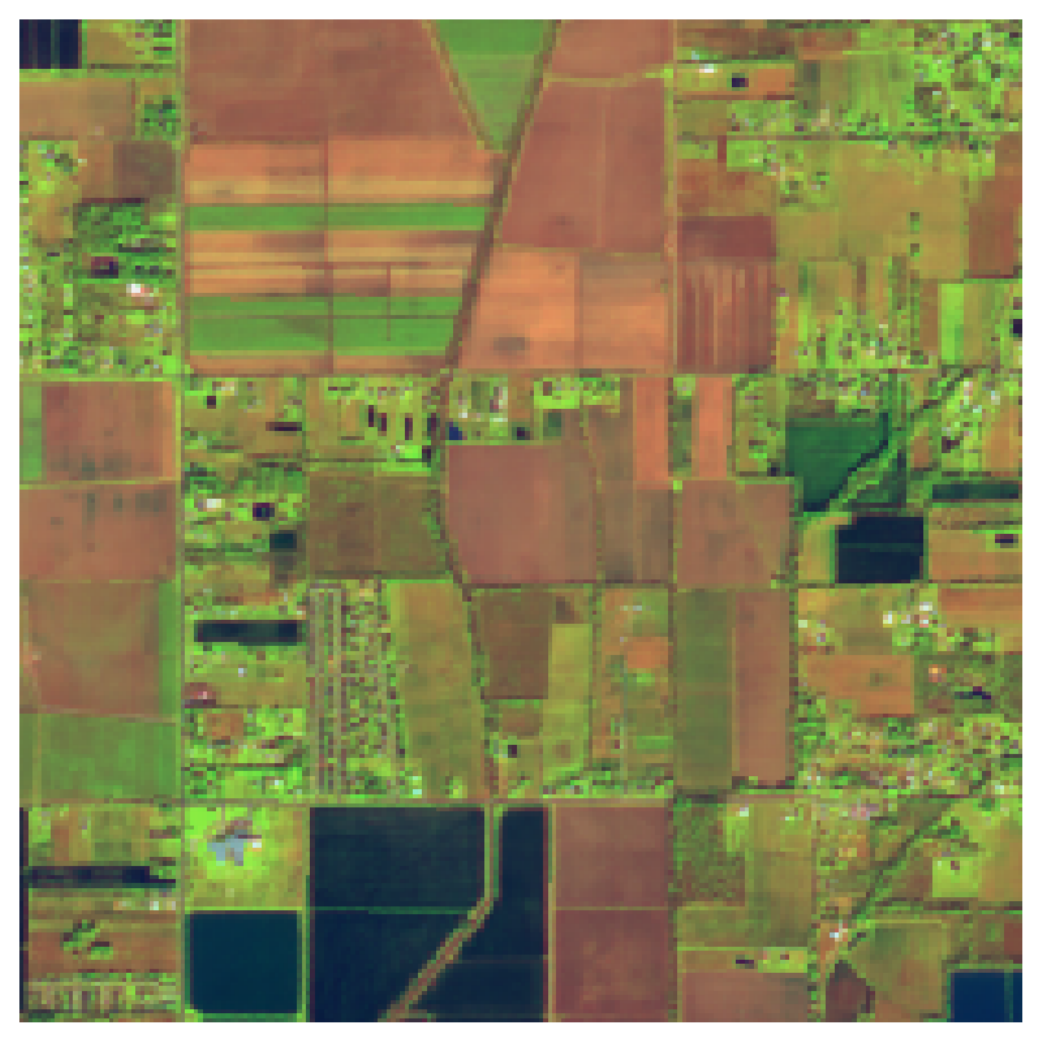}
      \caption{2022-12-15 \\ Original}
      \label{fig:sup-revist-ours-1215}
  \end{subfigure}
  \begin{subfigure}[t]{0.13\textwidth}
    \centering
    \includegraphics[width=\textwidth]{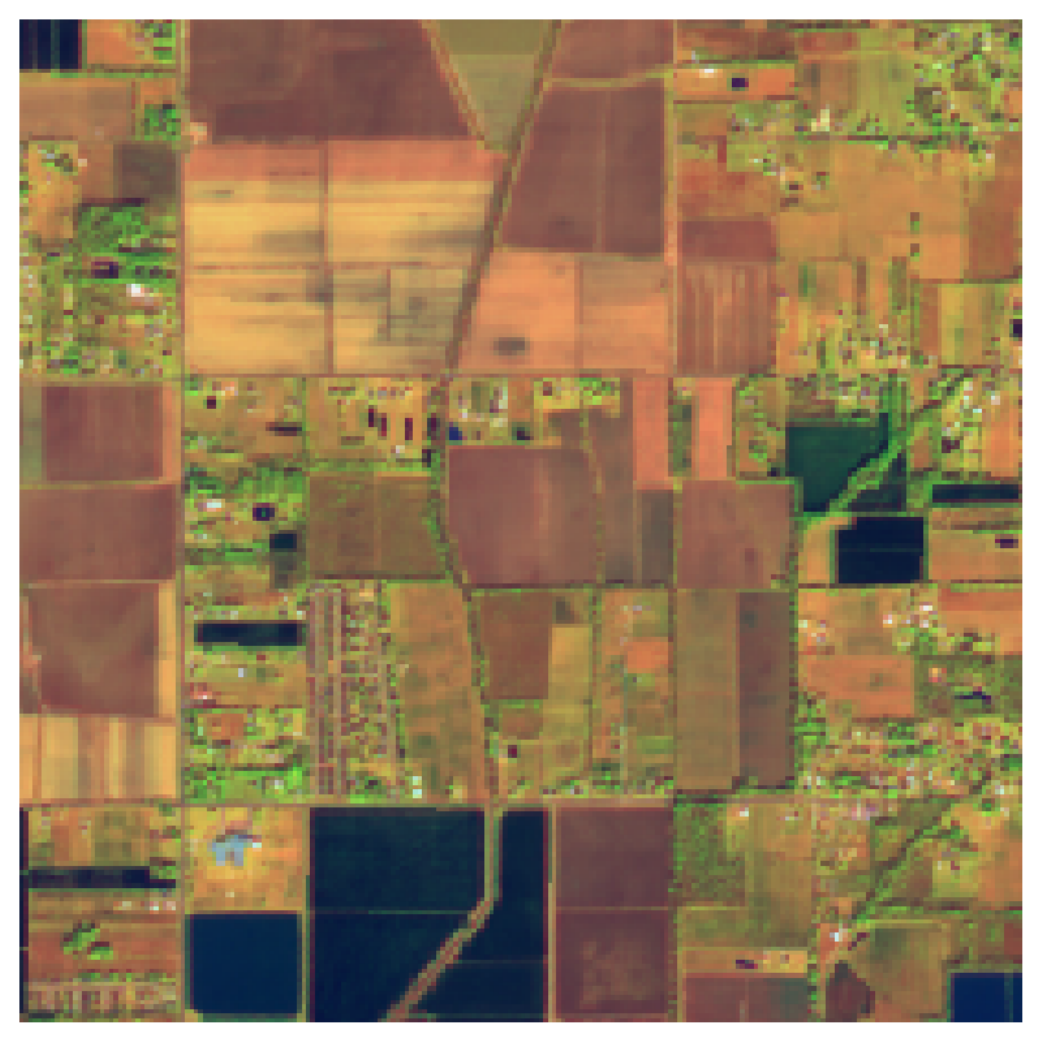}
    \caption{2023-01-01 \\ Original}
    \label{fig:sup-revist-ours-0101}
  \end{subfigure}
  \begin{subfigure}[t]{0.13\textwidth}
    \centering
    \includegraphics[width=\textwidth]{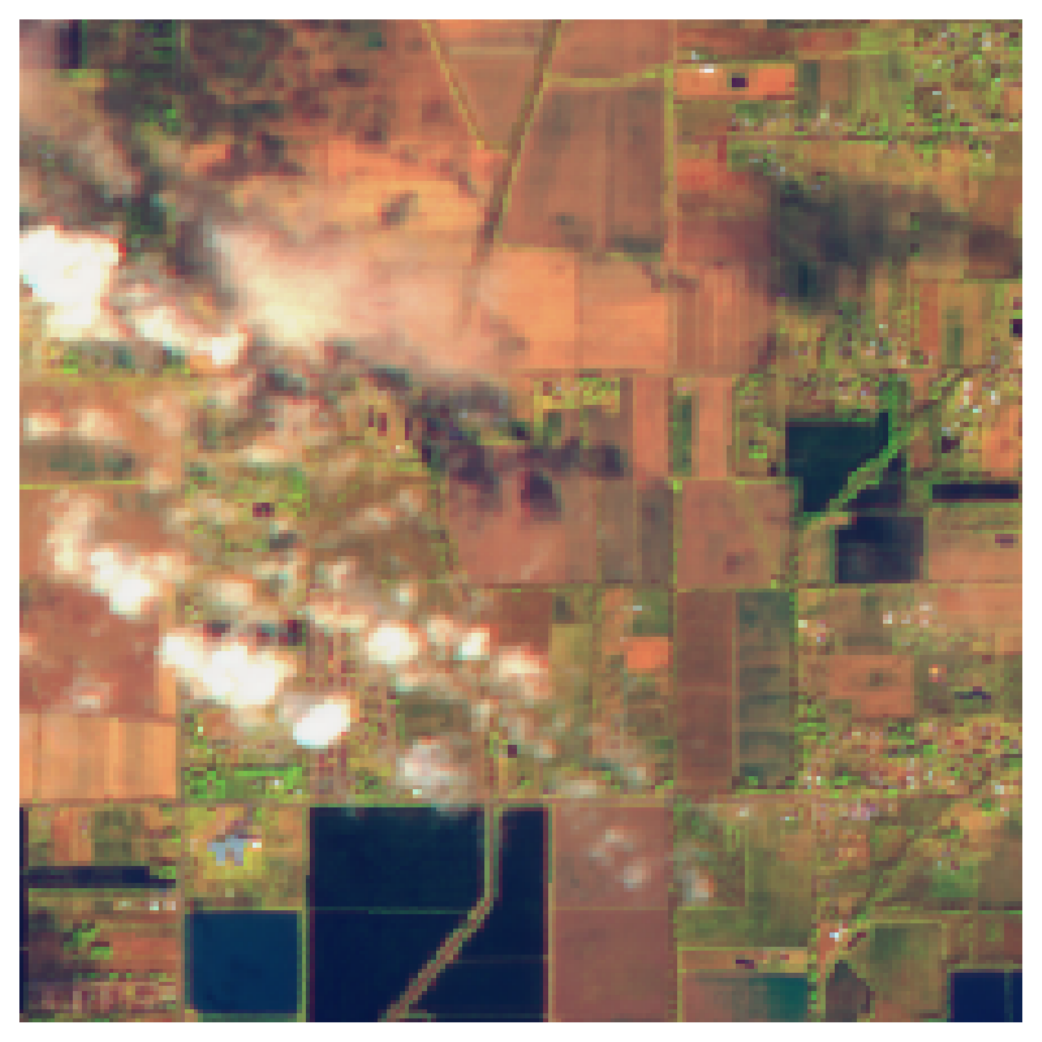}
    \caption{2023-01-15 \\ Original}
    \label{sup-revist-ours-0115}
  \end{subfigure}
  \begin{subfigure}[t]{0.13\textwidth}
    \centering
    \includegraphics[width=\textwidth]{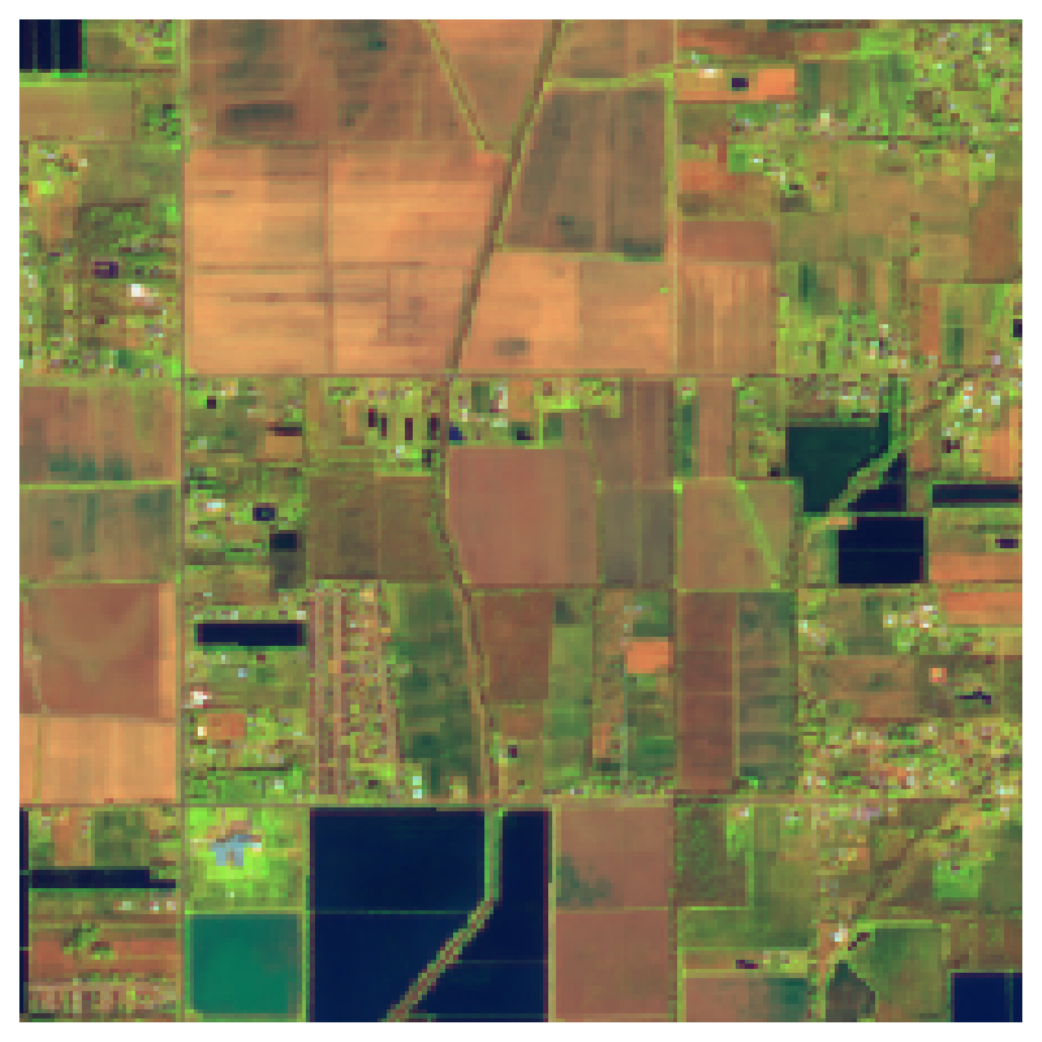}
    \caption{2023-02-01 \\ Original}
    \label{fig:sup-revist-ours-0201}
  \end{subfigure}
  \begin{subfigure}[t]{0.13\textwidth}
    \centering
    \includegraphics[width=\textwidth]{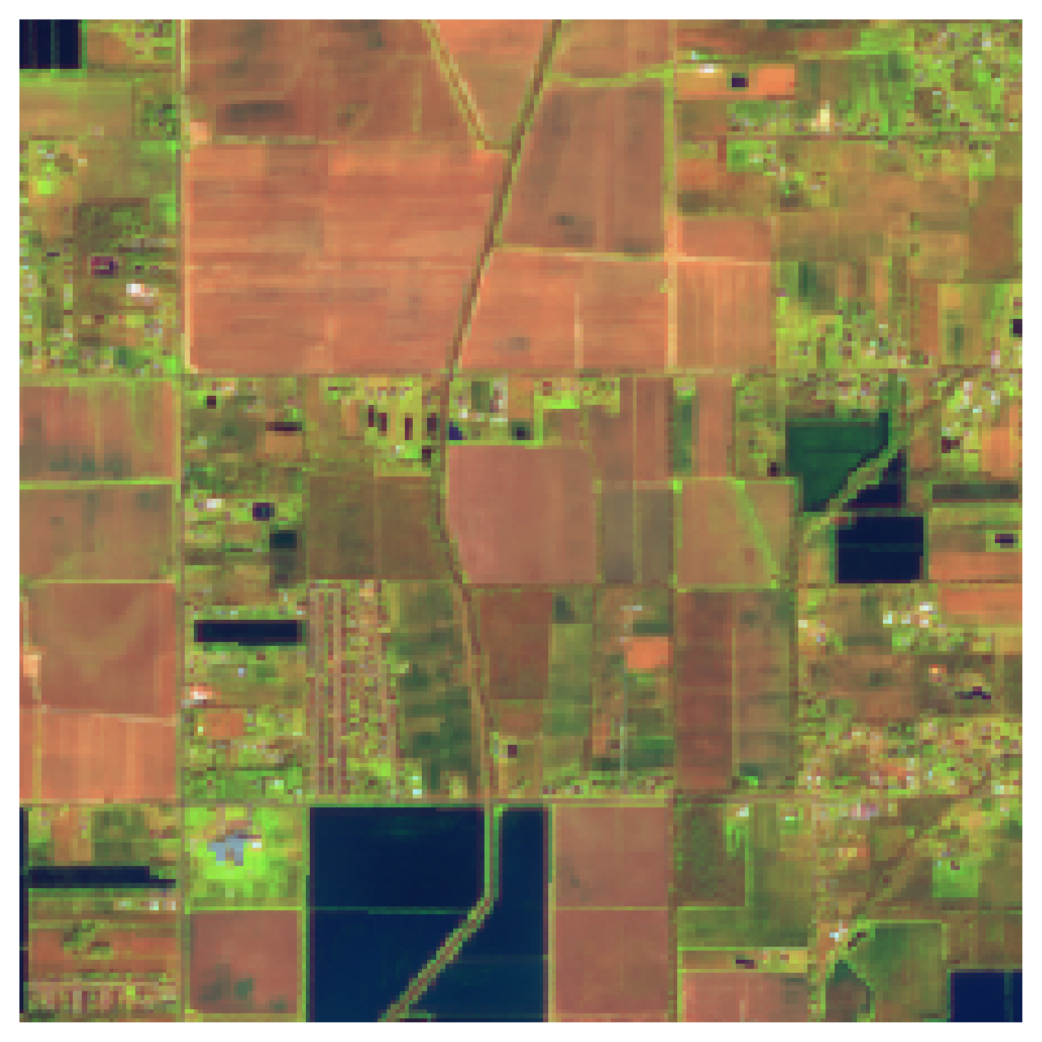}
    \caption{2023-02-15 \\ Original}
    \label{fig:sup-revist-ours-0215}
  \end{subfigure}
  \vspace{-0.5em}
  \caption{
    Examples of Sentinel-2 Imagery under 
    our revisit frequency of $14$ days 
    and our cloud coverage setting,
    with the revisit date listed below each image.
    We would like to highlight that there are no duplicate satellite images observed.
  }
  \label{fig:sup-revist-ours}
\end{figure*}

This section supplements the main paper by
demonstrating the necessity and importance of our cloud coverage setting (\ie\ $\leq 20 \%$)
and revisit frequency (\ie\ $14$ days) for Sentinel-2 Imagery.
Figures~\ref{fig:sup-w-cloud} and \ref{fig:sup-wo-cloud}
present examples of Sentinel-2 Imagery 
under the original revisit frequency of $5$ days 
with and without our cloud coverage setting, respectively.
Figure~\ref{fig:sup-revist-ours} illustrates satellites images 
under our revisit frequency of $14$ days and our cloud coverage setting (\ie\ $\leq 20 \%$).

From Figure~\ref{fig:sup-w-cloud},
we observed that the cloud coverage may significantly impair
the quality of Sentinel-2 Imagery 
(see Figures \ref{fig:sup-w-cloud-1206}, \ref{fig:sup-w-cloud-1216}, and \ref{fig:sup-w-cloud-1221}). 
Worse still, the extreme cases of cloud coverage 
(refer to Figures \ref{fig:sup-w-cloud-1216} and \ref{fig:sup-w-cloud-1221})
degrade satellite images into noisy representations.
This demonstrates the significance of our cloud coverage setting
for discarding low-quality satellite images.
Unfortunately, under the original sentinel-2 revisit frequency of $5$ days,
our cloud coverage setting would result in a large proportion of duplicate satellite images,
\eg\ $50 \%$ (\ie\ $3$ out of $6$ satellite images) as depicted in Figure \ref{fig:sup-wo-cloud}.
%
%
This is because if the cloud coverage in our requested revisit day exceeds $20 \%$,
Processing API~\cite{process-api} will download the most recent available satellite images,
whose cloud coverage satisfies our condition (\ie\ $\leq 20\%$).
In sharp contrast,
extending the revisit frequency from 5 days to 14 days 
markedly decreases the occurrence of duplicate satellite images.
For example, there are no duplicate satellite images observed in Figure~\ref{fig:sup-revist-ours}.
Hence, our revisit frequency of $14$ days for Sentinel-2 Imagery is necessary 
as it can significantly improve storage and training efficiency.

\begin{figure*}   [!t] 
  \centering
  \captionsetup[subfigure]{justification=centering}
  \begin{subfigure}[t]{0.25\textwidth}
      \centering
      \includegraphics[width=\textwidth]{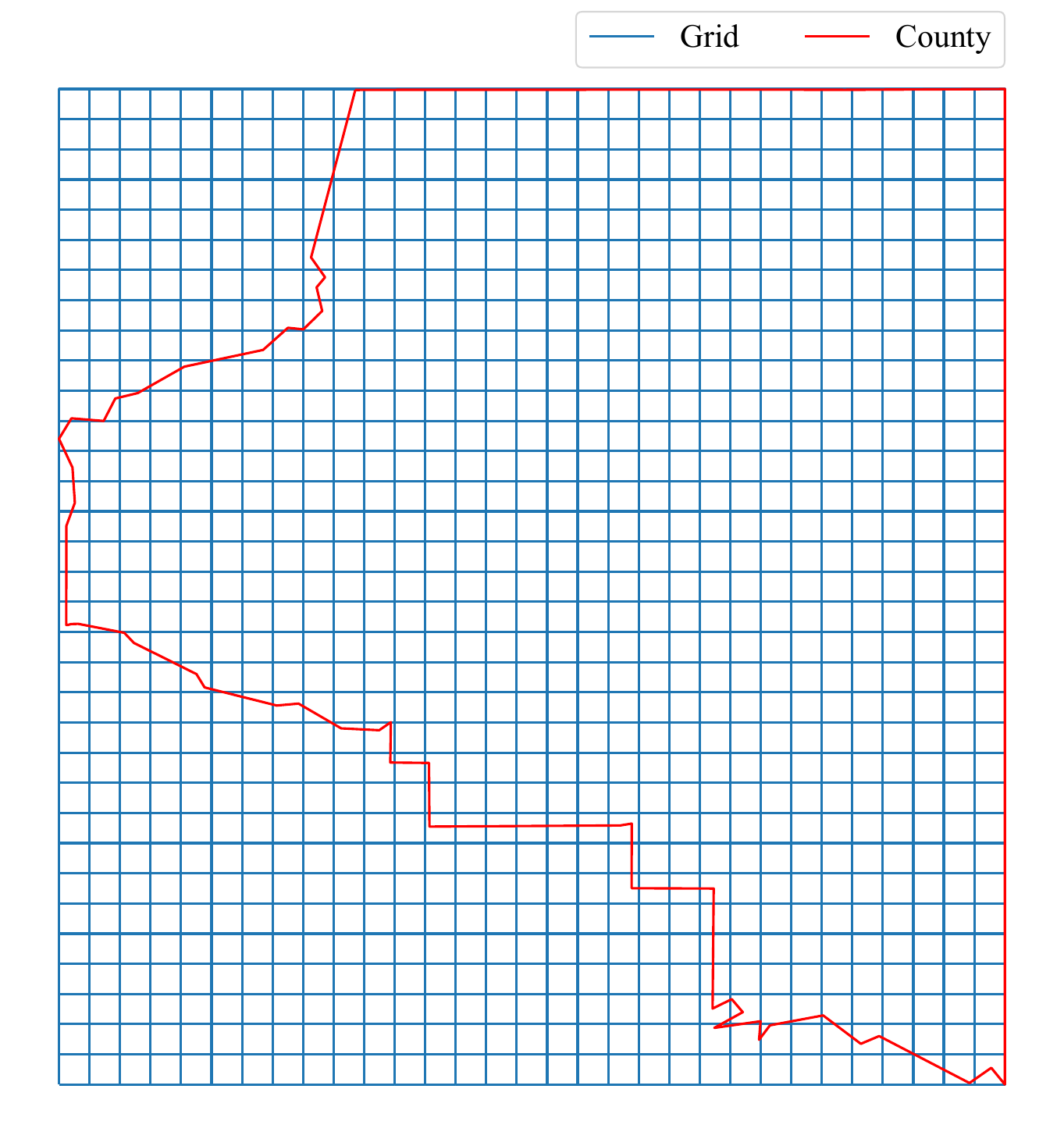}
      \caption{Naive solution}
      \label{fig:dataset-grid-wo-mask}
  \end{subfigure}
  \quad
  \begin{subfigure}[t]{0.25\textwidth}
      \centering
      \includegraphics[width=\textwidth]{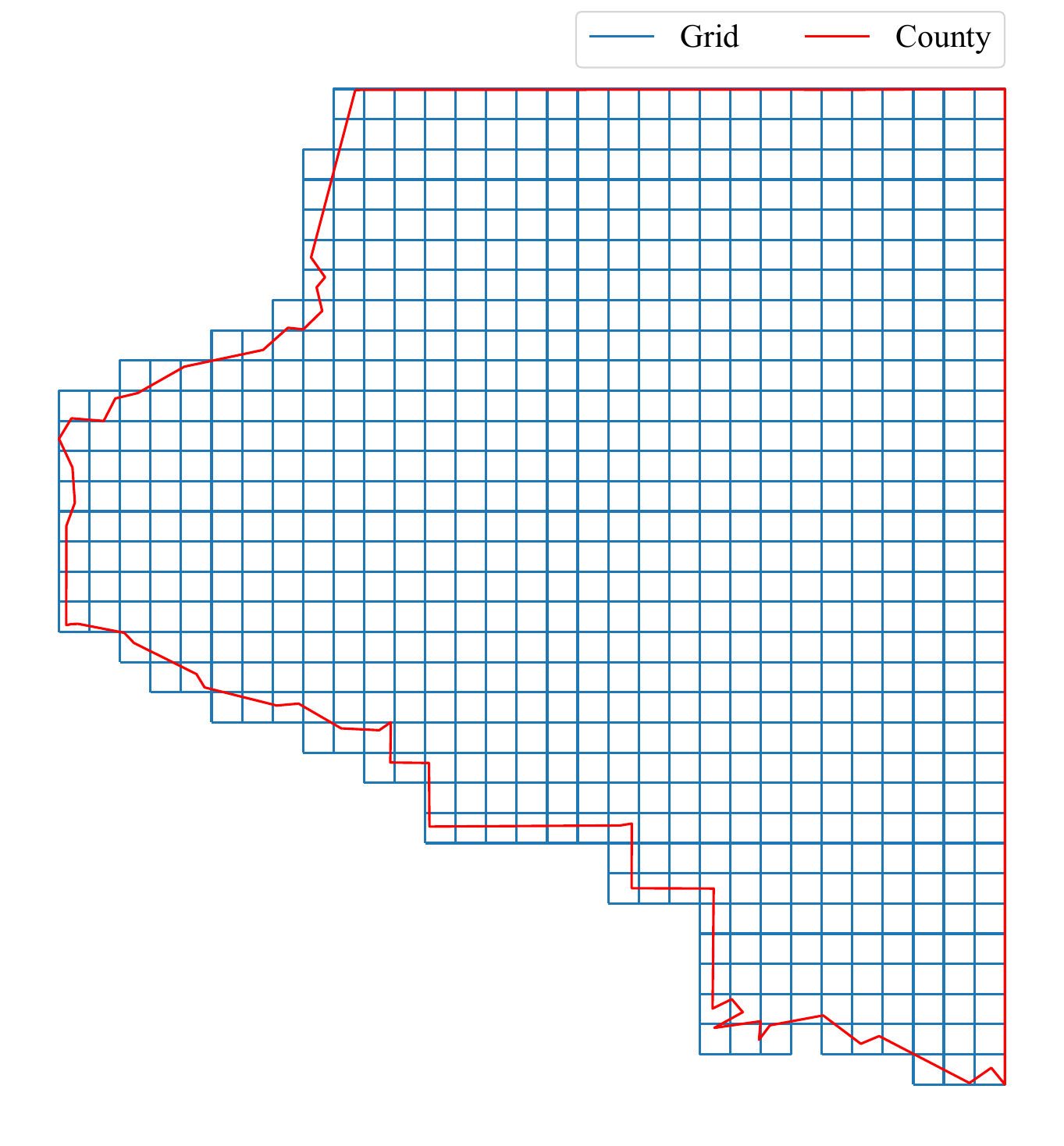}
      \caption{Our solution}
      \label{fig:dataset-grid-w-mask}
  \end{subfigure}
  \vspace{-0.5em}
  \caption{
    Difference between the naive solution and our solution. 
    (a) The naive solution leads to a significant number of grids falling outside the county's polygon.
    (b) By using our solution, the boundaries of grids (\ie\ the blue line) 
    align perfectly with the county's boundary (\ie\ the red line).
  }
  \label{fig:dataset-gird-masking}
  \vspace{-0.5 em}
\end{figure*}

\subsection{County Partitioning}
\label{sup:county-partition}

In our main paper, 
we have introduced partitioning one county into
multiple high-spatial-resolution grids for precise agricultural tracking.
Here, we provide the details for such a partition.
A naive way to achieve this is to expand a county's geographic boundary 
to a rectangle area by using its maximal and minimal latitude and longitude,
and then evenly divide such a rectangle area into multiple grids.
Unfortunately, such a partition solution may result in a large number of grids outside the county polygon 
for some large counties (see Figure~\ref{fig:dataset-grid-wo-mask}).
To handle this matter,
we develop a novel solution by dropping the grids outside the county's boundary
(see Figure~\ref{fig:dataset-grid-w-mask}).
Compared to the naive solution, our solution enjoys two advantages.
First, it can significantly reduce the disk space storage size.
Take Coconino County in Arizona for example, by employing our solution,
its total number of grids degrades from $1023$ to $729$,
which is $0.71$x less than that from the naive solution.
Second, our solution can evade the negative effect
incurred by regions outside the county's boundary on crop yield predictions.

\subsection{Spatial and Temporal Alignment of Our CropNet Dataset}
\label{sup:json-file}

Here, we present an example of our JSON configuration file (see Figure~\ref{list:config})
for one U.S. county (\ie\ Baldwin in Alabama),
to show how satellite images from Sentinel-2 Imagery, 
daily and monthly weather parameters from the WRF-HRRR Computed Dataset, 
and the crop information from USDA Crop Dataset,
are spatially and temporally aligned.
As presented in Figure~\ref{list:config}, 
``data.sentinel'' and ``data.HRRR.short\_term'' 
respectively represent satellite images and daily meteorological parameters during the crop growing season,
``data.HRRR.long\_term'' indicates monthly weather conditions from previous $5$ years,
and ``data.USDA'' provides the crop information for the county.
Meanwhile, ``FIPS'' and ``year'' respectively indicate the unique FIPS code 
and the year for the growing season, 
enabling us to obtain the data for our targeted county in a specific year.
In summary, the JSON configuration file allows us to 
retrieve all three modalities of data over the time and region of interest.

\begin{figure*} [!thp]
        \centering
        \begin{tabular}{c}
\begin{lstlisting}[language=Python, linewidth=12.5cm, label={list:config}]
{
    "FIPS":"01003",
    "year":2022,
    "county":"BALDWIN",
    "state":"AL",
    "county_ansi":"003",
    "state_ansi":"01",
    "data":{
        "HRRR":{
            "short_term":[
                "HRRR/data/2022/AL/HRRR_01_AL_2022-04.csv",
                "HRRR/data/2022/AL/HRRR_01_AL_2022-05.csv",
                "HRRR/data/2022/AL/HRRR_01_AL_2022-06.csv",
                "HRRR/data/2022/AL/HRRR_01_AL_2022-07.csv",
                "HRRR/data/2022/AL/HRRR_01_AL_2022-08.csv",
                "HRRR/data/2022/AL/HRRR_01_AL_2022-09.csv"
            ],
            "long_term":[
                [
                    "HRRR/data/2021/AL/HRRR_01_AL_2021-01.csv",
                    "HRRR/data/2021/AL/HRRR_01_AL_2021-02.csv",
                    "HRRR/data/2021/AL/HRRR_01_AL_2021-03.csv",
                    "HRRR/data/2021/AL/HRRR_01_AL_2021-04.csv",
                    "HRRR/data/2021/AL/HRRR_01_AL_2021-05.csv",
                    "HRRR/data/2021/AL/HRRR_01_AL_2021-06.csv",
                    "HRRR/data/2021/AL/HRRR_01_AL_2021-07.csv",
                    "HRRR/data/2021/AL/HRRR_01_AL_2021-08.csv",
                    "HRRR/data/2021/AL/HRRR_01_AL_2021-09.csv",
                    "HRRR/data/2021/AL/HRRR_01_AL_2021-10.csv",
                    "HRRR/data/2021/AL/HRRR_01_AL_2021-11.csv",
                    "HRRR/data/2021/AL/HRRR_01_AL_2021-12.csv"
                ],
                # The remaining years are hidden for conserving space
                ...
            ]
        },
        "USDA":"USDA/data/Soybeans/2022/USDA_Soybeans_County_2022.csv",
        "sentinel":[
            "Sentinel/data/AG/2022/AL/Agriculture_01_AL_2022-04-01_2022-06-30.h5",
            "Sentinel/data/AG/2022/AL/Agriculture_01_AL_2022-07-01_2022-09-30.h5"
        ]
    }
}
\end{lstlisting}
    \end{tabular}
    \caption{Example of our JSON configuration file.}
    \label{list:config}
    \vspace{-1.0 em}        
\end{figure*}


\section{Supporting Experimental Settings} 
\label{sup:exp-setup}



\par\smallskip\noindent
{\bf CropNet Data.}
Due to the limited computational resources, 
we are unable to conduct experiments across the entire United States.
Consequently, we extract the data with respect to five U.S. states, 
\ie\ Illinois (IL),  Iowa (IA), Louisiana (LA), Mississippi (MS), and New York (NY),
to exhibit the applicability of our crafted CropNet dataset
for county-level crop yield predictions.
Specifically, two of these states (\ie\ IA and IL) serve as representatives of the Midwest region,
two others (\ie\ LA and MS) represent the Southeastern region, 
and the fifth state (\ie\ NY) represents the Northeastern area.
Four of the most popular crops are studied in this work, \ie\ corn, cotton, soybeans, and winter wheat. 
For each crop, we take the aligned  Sentinel-2 Imagery and the daily data in the WRF-HRRR Computed Dataset
during growing seasons in our CropNet dataset, respectively for precise agricultural tracking
and for capturing the impact of growing season weather variations on crop growth.
Meanwhile, the monthly meteorological parameters from the previous $5$ years are utilized 
for monitoring and quantifying the influence of climate change on crop yields. 

\end{document}